\documentclass[10pt,twocolumn,letterpaper]{article}
\pdfoutput=1
\usepackage{iccv}
\usepackage{times}
\usepackage{epsfig}
\usepackage{graphicx}
\usepackage{amsmath}
\usepackage{amssymb}
\usepackage{multicol}
\usepackage{hhline}
\usepackage{color}
\usepackage{xcolor}
\usepackage{caption}
\usepackage{subcaption}
\usepackage{soul}
\usepackage[ruled,linesnumbered]{algorithm2e}

\iccvfinalcopy % *** Uncomment this line for the final submission

\newcommand{\figref}[1]{Figure~\ref{#1}}

\begin{document}

%For conv kernels 5x5, 3x3 etc
\newcommand{\kernel}[1]{${{#1}\!\!\times\!\!{#1}}$}

%%%%%%%%% TITLE
\title{AutoDispNet: Improving Disparity Estimation With AutoML}

\author{
Tonmoy Saikia
\and
Yassine Marrakchi\\
\and
Arber Zela\\
\and
Frank Hutter\\
\and
Thomas Brox\\
\and \\
University of Freiburg\\
Germany\\
{\tt\small \{saikiat, marrakch, zelaa, fh, brox\}@cs.uni-freiburg.de}
%\and
}
\maketitle

\begin{abstract}
Much research work in computer vision is being spent on optimizing existing network architectures to obtain a few more percentage points on benchmarks. Recent AutoML approaches promise to relieve us from this effort. However, they are mainly designed for comparatively small-scale classification tasks.
In this work, we show how to use and extend existing AutoML techniques to efficiently optimize large-scale U-Net-like encoder-decoder architectures. In particular, we leverage gradient-based neural architecture search and Bayesian optimization for hyperparameter search. The resulting optimization does not require a large-scale compute cluster. We show results on disparity estimation that clearly outperform the manually optimized baseline and reach state-of-the-art performance.
\end{abstract}

\section{Introduction}
Compared to the state of computer vision 20 years ago, machine learning has enabled more generic methodologies that can be applied to various tasks rather than a single toy problem. A convolutional neural network can be trained on all sorts of classification problems, and a convolutional encoder-decoder network with skip connections can be set up for a large selection of high-resolution computer vision tasks, such as semantic segmentation, optical flow, super-resolution, and depth estimation, to name just a few. With this generic methodology in place, why are there more than 5000 submissions to each computer vision conference? What do they contribute? 

\begin{figure}
\begin{center}
\includegraphics[width=0.9\linewidth]{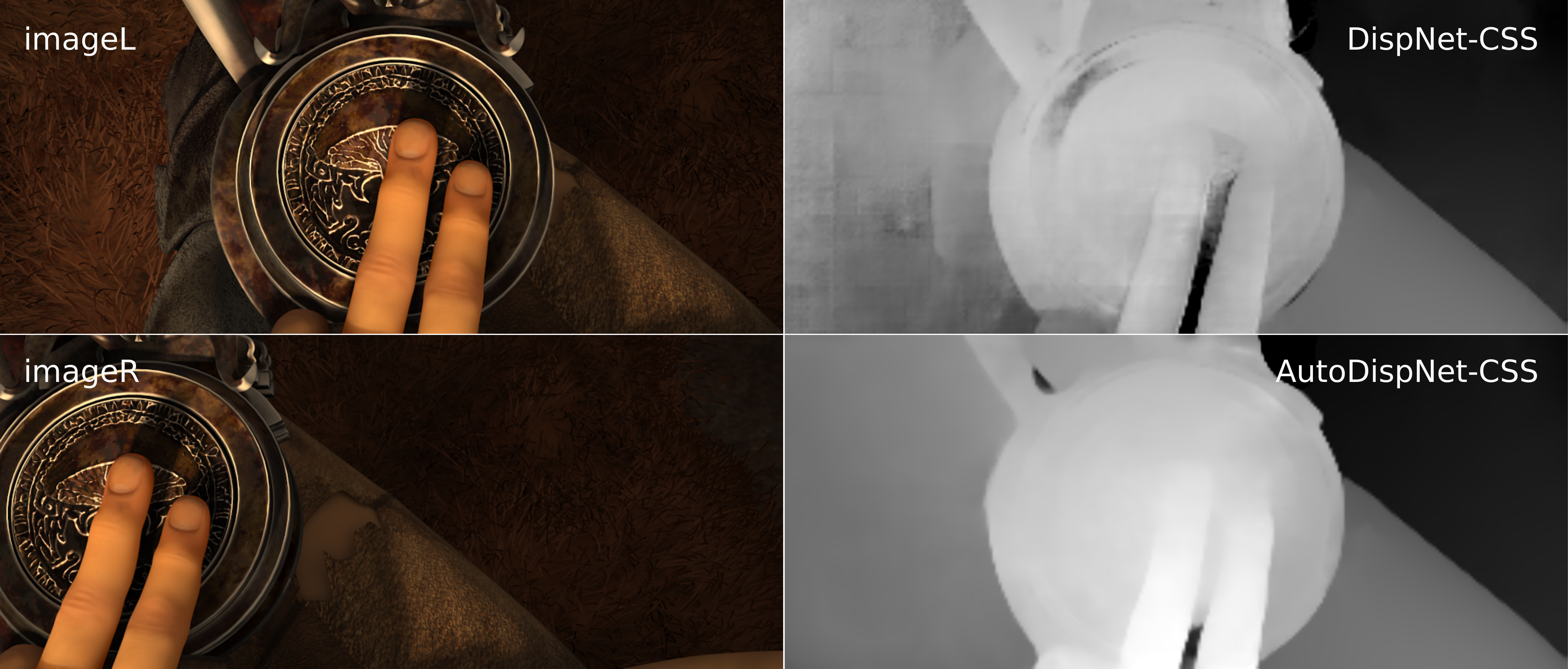}
\end{center}
\caption{We present AutoDispNet-CSS, a disparity estimation network built using state-of-the-art AutoML techniques which significantly improves over the manually tuned DispNet-CSS architecture.} 
\end{figure}

In practice, the methodology is not as generic as it looks at first glance. While a standard encoder-decoder network may give a reasonable solution for all these problems, results can be improved significantly by tweaking the details of this design: both the detailed architecture and several training hyperparameters, such as the learning rate and weight decay. For example, in the context of disparity estimation, manually optimizing the architecture of the original DispNet~\cite{dn} halved the error~\cite{dn3}. Other works on disparity estimation found other tweaks that also improved the accuracy~\cite{iresnet, psmnet,edge_stereo}. While effective, this common practice of manual architecture and parameter search contradicts the paradigm of machine learning, namely to replace manual optimization by numeric optimization. %\tcf{I'd rephrase ``numeric optimization'' to ``data-driven automatic optimization''.} 

AutoML~\cite{automl_book} in general and automated neural architecture search (NAS~\cite{elsken_neural_2018}) in particular promise to relieve us from the manual tweaking effort. In principle, an independent validation set is enough to optimize the architecture and the hyperparameters of the learning method. Unfortunately, many of these AutoML methods have extreme computational demands. For this reason, they have been mostly applied to rather small-scale classification tasks, preferably on CIFAR, where a single network can be fully trained within a few hours. Even on such small tasks, some approaches report hundreds of GPU days to finish optimization. For large-scale encoder-decoder networks, such as DispNet, this is prohibitive. 

However, there are also more efficient AutoML approaches. Although they have not yet been applied to encoder-decoder architectures, they have the potential to do so. One of them, on which we will build in this paper, is DARTS~\cite{darts}. Its main idea is to have a large network that includes all architectural choices and to select the best parts of this network by optimization. This can be relaxed to a continuous optimization problem, which, together with the regular network training, leads to a bilevel optimization problem. Thanks to its gradient based optimization, DARTS is very efficient. 
However, DARTS only allows the optimization of the architecture but not the training hyperparameters. 

For the latter, we propose to use an efficient black-box optimization method called BOHB~\cite{Falkner18}, which builds on an efficient variant of Bayesian optimization for informed sampling of the search space. While it is somewhat more costly than DARTS, it is also entirely flexible with regard to the hyperparameter search space. We suggest to run BOHB on the architecture optimized by DARTS to train it with optimal hyperparameters.  

We compare the optimized network to the already well-tweaked version of DispNet~\cite{dn3} to investigate who is more successful in tweaking: the student or the numerical optimization procedure.

\section{Related work}

Encoder-decoder architectures have led to substantial improvements in several computer vision tasks like semantic segmentation~\cite{Long15,Ronneberger15,Badrinarayanan17,He14,Chen17,ChenL18} and flow estimation~\cite{Dosovitskiy15,Ilg17,dn3,ranjan17,pwc}. Pioneering works which apply learning to disparity estimation consist of extending classical methods like SGM~\cite{Hirschmuller08} with metrics learned by CNNs~\cite{Zbontar16,Luo16,SekiP17}. The first end-to-end network for disparity estimation is DispNet~\cite{dn}, which builds on FlowNetC~\cite{Dosovitskiy15}. Based on rectified stereo images, a correlation layer computes a cost volume which is further processed by the network.  ~\cite{dn3} and~\cite{Pang17} expand DispNetC for much better performance. The extensions consist mainly of stacking multiple networks and connecting them in a residual fashion. These networks share the encoder-decoder architecture. The first module, the encoder, extracts high-level information by gradually downsampling the feature maps while the decoder progressively produces outputs at increasing resolutions. 

To reduce the effort dedicated to designing neural networks, neural architecture search (NAS) has been an active area of research in the last few years~\cite{elsken_neural_2018}. Early attempts train a recurrent neural network that acts as a meta-controller using reinforcement learning techniques~\cite{Baker16,Zoph16}. It learns to generate sequences encoding potential architectures by exploring a predefined search space. The same strategy is adopted in many follow-up works~\cite{Bello17,Cai17,Cai18a,Tan18,Zhao17,Zhong17,Zoph17}.  Alternatively, a set of works rely on evolutionary algorithms~\cite{Stanley02,Liu17,Miikkulainen17,Real18,Real17,Xie17}. The best architecture is extracted by iteratively mutating a population of candidate architectures. Unfortunately, both strategies require hundreds to thousands of GPU days. 
This restricts their use to rather small networks, and research progress is limited by availability of large compute clusters.  

Speed-up techniques like hypernetworks, network morphisms and shared weights lead to substantial reduction of the search cost.
Hypernetworks~\cite{Brock17,Zhang19} generate weights for candidate networks and evaluate them without training them until convergence. Network morphisms~\cite{Cai18, Cai18a, Elsken17, Elsken19} make use of the previous learned weights to initialize new candidate architectures, thereby speeding up the performance estimation procedure. Sharing weights~\cite{Pham18} among potential networks decreases the search time by two orders of magnitude. Multi-fidelity optimization has also been employed in NAS~\cite{baker_accelerating_2017, Falkner18, li17, Zela18} by exploiting partial training of architectures at the cost of noisy evaluations. Alternatively, some works~\cite{Bender18,Cai19,darts} redesign the optimization problem by training a large graph containing all candidate architectures. In~\cite{Bender18}, sub-networks are probabilistically sampled and trained for a predefined number of iterations. Orthogonally, relaxations make architectural decisions like branching patterns~\cite{Ahmed18} and number of channels per layer~\cite{Saxena16} learnable via gradient descent. In case of DARTS~\cite{darts},  real-valued architecture parameters are jointly trained with weight parameters via standard gradient descent. Cai \etal~\cite{Cai19} 
propose an memory efficient implementation similar to DARTS by adding path binarization, while \cite{Xie18} sample from a set of one-hot random variables encoding the architecture search space and leverage the gradient information for architectural updates by relaxing the architecture distribution with a concrete distribution~\cite{Maddison17}.
Despite the diversity of NAS approaches for image classification and object detection, the extension to dense prediction tasks remains restricted. To apply NAS to semantic segmentation, Chen et al.~\cite{Chen18} restrict the search to the small pyramid pooling component of the network and occupy $370$ GPUs for a whole week. In a concurrent work, Liu~\etal~\cite{autodeeplab} also leverage DARTS to find an optimal architecture for semantic segmentation with reduced search cost. However, their approach does not handle skip-connections for U-Net like architectures.

\section{Hyperparameter search}
\label{sec:bohb}
Optimizing hyperparameters for dense prediction tasks with vanilla hyperparameter optimization (HPO)~\cite{bergstra13, bergstra11, Hutter13b, snoek12, snoek15, feurer_hyperparameter_2018} is computationally expensive. Alternatively, we use a state-of-the-art HPO method named BOHB \cite{Falkner18} which combines the benefits of Bayesian optimization \cite{Shahriari16} and Hyperband \cite{li17}, a multi-armed bandit strategy that dynamically allocates more resources to promising configurations.

BOHB uses cheap-to-evaluate approximations $\Tilde{f}(\cdot, b)$ of the objective function $f(\cdot)$ (e.g. validation error), where the so-called budget $b \in \lbrack b_{min}, b_{max} \rbrack$ determines the strength of the approximation. For $b = b_{max}$, we recover the true objective, i.e. $\Tilde{f}(\cdot, b_{max}) = f(\cdot)$. 
In our application, we use the number of training iterations as a budget to cut off evaluations of poorly-performing hyperparameters early, akin to approaches based on learning curve prediction~\cite{Domhan15,baker_accelerating_2017}.

Hyperband repeatedly calls the Successive Halving (SH) subroutine \cite{Jamieson16} to advance promising configurations evaluated on small budgets to larger ones. SH starts by evaluating a fixed number of configurations 
%determined by $\eta$, $b_{min}$ and $b_{max}$ 
on the cheapest budget $b_{min}$. 
After these evaluations, the best fraction of $\eta^{-1}$ of configurations (based on $\Tilde{f}(\cdot, b_{min})$) advance to the next budget $\eta \cdot b_{min}$; here, $\eta$ is a parameter set to 3 by default. This procedure repeats until reaching the most expensive budget $b_{max}$ with only a few configurations left to evaluate. 

While Hyperband selects configurations to evaluate uniformly at random, BOHB replaced this choice with Bayesian optimization. Specifically, it employs a multivariate kernel density estimator (KDE) to model the densities of the best and worst performing configurations and uses these KDEs to select promising points in the hyperparameter space to evaluate next.
More details about BOHB are included in the supplementary material. 
% ? 

\section{Differential architecture search}
\begin{figure*}
\begin{center}
\subcaptionbox{Normal/Reduction cell \label{fig:cell_norm_reduce_darts}}{\includegraphics[width=0.32\linewidth]{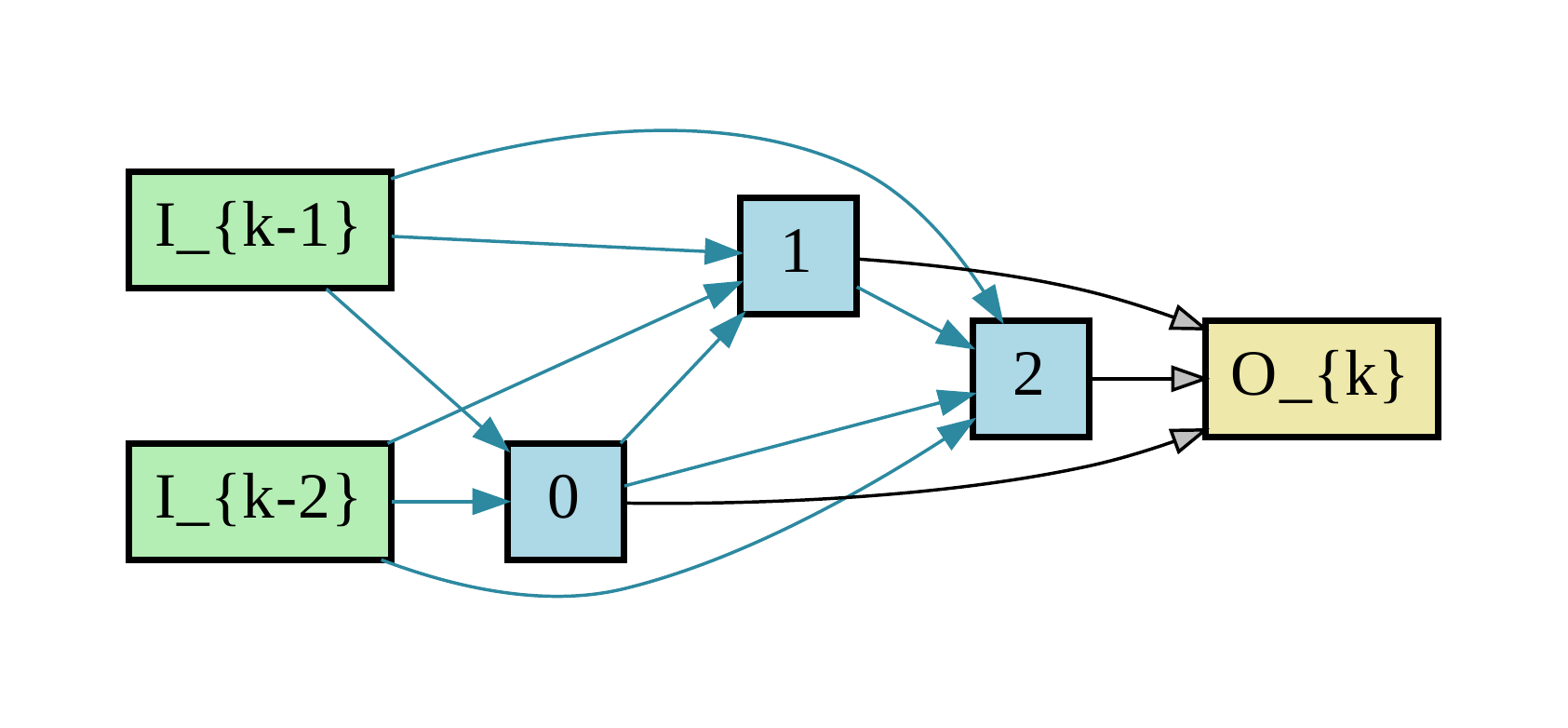}}%
%\hfill 
\subcaptionbox{Upsampling cell\label{fig:cell_upsampling}}{\includegraphics[width=0.45\linewidth]{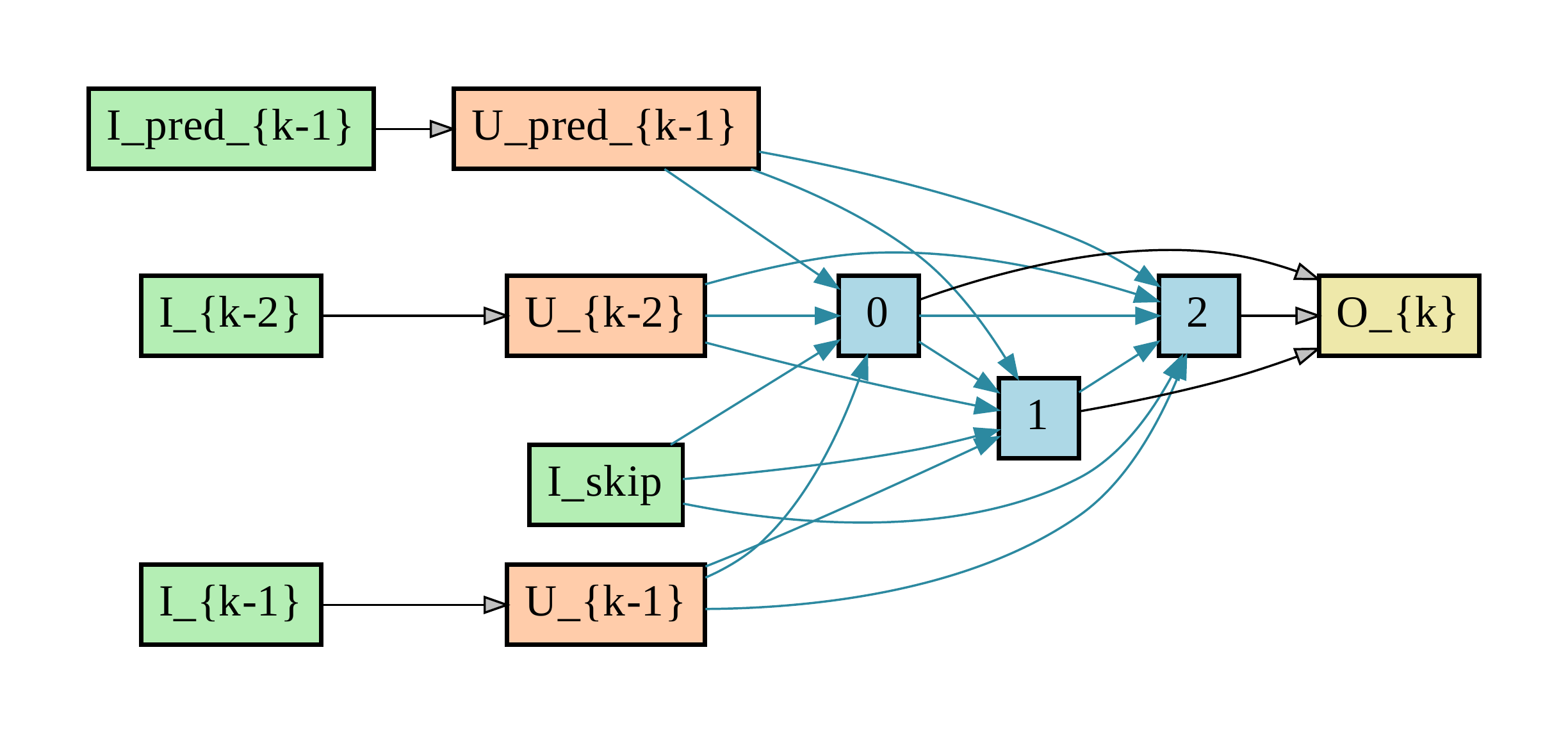}}%
\end{center}
\caption{Structure of search cells. In Figure \ref{fig:cell_norm_reduce_darts} we show the structure of a normal or reduction cell. An upsampling cell is shown in Figure \ref{fig:cell_upsampling}.
In both cases, input nodes are green, intermediate nodes are blue, output nodes are yellow. Upsampling nodes are marked as orange. A blue edge represents transformations done using mixed operations 
(see Section \ref{sec:mixed_ops} for more details). 
} 
\label{fig:cell_structure}
\end{figure*}

While BOHB can, in principle, also be used to optimize architectural parameters~\cite{Zela18,runge2018learning}, its performance degrades compared to gradient-based approaches as the dimensionality of the search space grows.
%Compared to gradient-based approaches, BOHB also has the disadvantage of having to
BOHB also evaluates different architectures from scratch rather than exploiting weight sharing, increasing the computational burden for neural architecture search of large-scale vision architectures to a prohibitive range.

Therefore, we tackle the neural architecture search not with BOHB, but rather use the gradient-based method DARTS~\cite{darts}. It combines weight sharing and first order optimization to speed up the architecture optimization by orders of magnitude compared to 
brute-force blackbox optimization methods, which can require thousands of GPU days \cite{Real18, Zoph17}. 
We propose to leave the costly architecture search to DARTS and then optimize important other hyperparameters that cannot be integrated into DARTS by BOHB in a post-hoc step. 

We review the components of the DARTS approach before we adapt DARTS to the context of full encoder-decoder architectures in Section~\ref{sec:dense_darts}.

\subsection{Search space}
\label{sec:search_space}
Similar to other architecture search methods \cite{Liu17,Real18,Zoph17}, DARTS optimizes relatively small, repetitive parts of the network architecture called cells. 
%structures depicted in Figure \ref{fig:cell_structure}a. 
Learned cells are stacked to generate the overall network architecture in a user defined fashion.

A cell is a directed acyclic graph (DAG) consisting of $N$ nodes. The nodes can be categorized into input, intermediate, and output nodes. 
Each node $x^{(i)}$ represents a feature map and each edge $(i,j)$ represents an operation $o^{(i,j)}$ which transforms $x^{(i)}$. 
DARTS assumes a cell to have two input nodes and one output node. The input nodes are the outputs of the two previous cells. 
The output node is obtained by concatenating the outputs of all intermediate nodes. The output of each intermediate node is computed as: 
\begin{equation} \label{eq1}
\begin{split}
x^{(j)} = \sum_{i<j}o^{(i,j)}(x^{(i)})\\
\end{split}
\end{equation}
where $o^{(i,j)} \in \mathcal{O}$. $\mathcal{O}$ is the set of all candidate operations. 

In DARTS, $\mathcal{O}$ consists of the following operations: skip connection, \kernel{3} average pooling, \kernel{3} max pooling, \kernel{3} and \kernel{5} depthwise separable convolutions, \kernel{3} and \kernel{5} dilated separable convolutions with dilation factor $2$. It also includes a special "zero" operation to indicate lack of connectivity between nodes. 

For classification tasks there are two cell types: a \emph{normal cell} with maintains the spatial resolution of the input and a \emph{reduction cell} which reduces the spatial resolution of the input by half. The structure of standard DARTS cell is shown in Figure \ref{fig:cell_norm_reduce_darts}. 

\subsection{Continuous relaxation}
\label{sec:mixed_ops}
To make the search space continuous, DARTS uses relaxation based on the softmax function.
A variable $\alpha_{o}^{(i,j) }\in \mathbb{R}$ is associated with each operation $o \in \mathcal{O} $ in the edge $(i,j)$ connecting nodes $i$ and $j$. The categorical choices in each edge $(i, j)$ are then relaxed by applying the softmax nonlinearity over the $\alpha_{o}^{(i,j) }$ for all possible operations $o \in \mathcal{O}$:
\begin{equation} \label{}
    \begin{split}
        S^{(i,j)}_o  = \frac{\exp(\alpha_o^{i,j})}{\sum_{o' \in \mathcal{O}}\exp(\alpha_{o'}^{i,j})}
    \end{split}
\end{equation}
We set $\bar{o}^{(i,j)}=\sum_{o\in\mathcal{O}} S^{(i,j)}_oo\left(x^{(i)}\right)$. This weighted average of $\left(x^{(i)}\right)$ is called "mixed operation" in the remainder of this work. Therefore, \eqref{eq1} becomes:
\begin{equation} \label{eq:node_output}
\begin{split}
x^{(j)} = \sum_{i<j}\bar{o}^{(i,j)}(x^{(i)})\\
\end{split}
\end{equation}
With this relaxation in place, the task of architecture search is equivalent to learning the set of continuous variables $\alpha = \{\alpha^{(i,j)}\}$, where $\alpha^{(i,j)}$  is a vector of dimension $|\mathcal{O}|$.

\subsection{Optimization}
\label{sec:search_opt}
Since the continuous relaxation makes the set of architecture variables $\alpha$ differentiable, we can efficiently optimize them using gradient descent. DARTS \cite{darts} proposed a first order and second order approximation. In this work, we focus on the first order approximation as the second order approximation is too costly for large architectures. 

To solve the bilevel optimization problem, the training data is split into two disjoint subsets  $\mathcal{D}_\text{train}$ and $\mathcal{D}_\text{val}$.
The network parameters $w$ and architecture parameters $\alpha$ are optimized in an alternating fashion on $\mathcal{D}_\text{train}$ and $\mathcal{D}_\text{val}$, respectively,  until convergence. The optimization is carried out on a search network built using stacked normal and reduction cells. 

\subsection{Architecture discretization}
\label{sec:search_and_extraction}

After training the search network to convergence, a cell structure is extracted by discretizing the continuous variables. This is achieved by retaining the top-$k$ strongest operations from all non-zero operations coming from previous nodes. The strength of an edge $(i,j)$ is set to: 
\begin{equation} \label{}
    \begin{split}
        \max_{o \in \mathcal{O}, o\neq \textrm{zero}} S^{(i,j)}_o
    \end{split}
\end{equation}
The extracted cells are then stacked to form a deeper network and retrained for evaluation.

\section{DARTS for dense prediction}
\label{sec:dense_darts}
\begin{figure*}
\begin{center}
\subcaptionbox{Search network\label{fig:search_net}}{\includegraphics[width=0.72\linewidth]{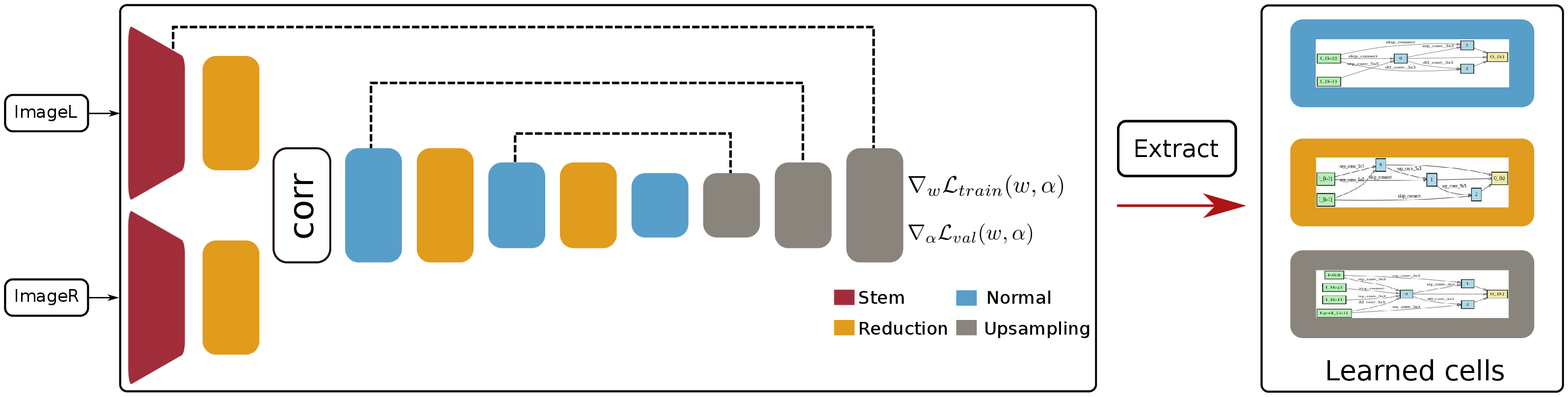}}
\\
\subcaptionbox{AutoDispNet-CSS\label{fig:autodispnet}}{\includegraphics[width=0.9\linewidth]{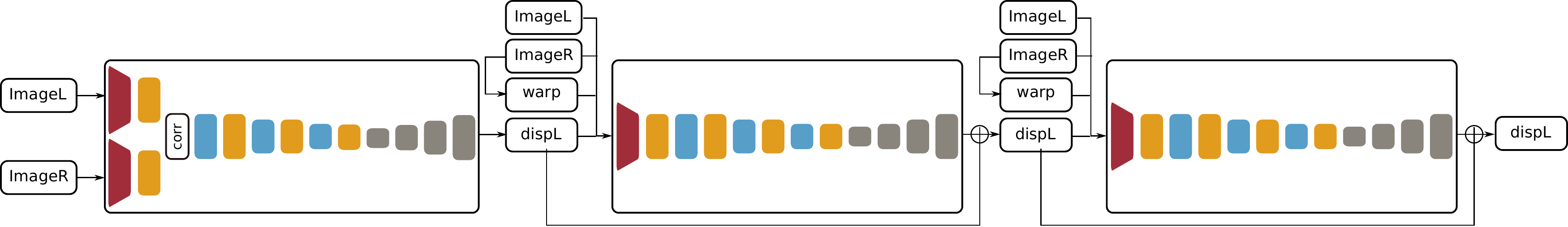}}
\end{center}
\caption{Dense-DARTS for disparity estimation.  The search network used to learn cells for disparity estimation is shown in Figure \ref{fig:search_net}. Three types for cells are learned: normal, reduction and upsampling. The stem cells are simple convolutional layers with a fixed structure. It also contains a correlation layer like a standard DispNetC~\cite{dn}. Skip connections from encoder to decoder are denoted by the dashed lines. After training, the three cell structures are extracted as described in Section \ref{sec:search_and_extraction}. Using the extracted cells, a final network (Figure \ref{fig:autodispnet}) is assembled using the CSS structure introduced in~\cite{dn3}.}
\label{fig:search_arch}
\end{figure*}

Dense prediction tasks involve mapping a feature representation in the encoder to predictions of larger spatial resolution using a decoder.
Therefore, to apply DARTS for disparity estimation we need to extend the architecture search space such that it can support an upsampling transformation.
This extension of the search space should be expressive enough to encompass common deep learning best-practices and at the same time have enough flexibility to learn new upsampling transformations. In this section, we describe our search space and then present a search network which allows us to learn architectural cells for encoder-decoder networks.

\textbf{Upsampling layers.} Typically, the decoder unit of encoder-decoder networks~\cite{dn,Dosovitskiy15,Ronneberger15} consists of upsampling layers which increase the spatial resolution. The most commonly used upsampling layers are transposed convolutions. Another common approach is to use billinear interpolation for upsampling followed by convolutional operations. A decoder usually has multiple decoding stages, each of which increases the spatial resolution by a factor of $2$. The number of stages depends on the downsampling factor of the encoder.

\textbf{Skip connections and multi-scale refinement.} Skip connections were introduced in encoder-decoder networks by ~\cite{Long15}. They help preserve fine details in the predictions. This is achieved by concatenating the upsampled features with a feature map of the same resolution from the encoder. A coarser prediction (from the previous decoding stage), if available, is also concatenated to facilitate feature reuse. The concatenated features are then processed by convolutions to generate refined predictions. These techniques are standard for encoder-decoder networks for flow and disparity estimation~\cite{dn, Dosovitskiy15}.

\textbf{Upsampling cell.} Several hand-designed encoder-decoder architectures have emerged \cite{Islam18,shuffle_seg,ZhangZ18, densenet_flow} which incorporate the above architecture design choices.
Typically such methods propose decoding modules which apply architectural blocks (ShuffleNet~\cite{shufflenet}, DenseNet~\cite{densenet} block, etc). However, the generic design choice of having skip connections and multi-scale refinement still remains useful in such cases. In this work, we replace such an architectural block in the decoder by a learned upsampling cell. The same DAG-based formulation for normal and reduction cells (see Section \ref{sec:search_space}), can  be used to define an upsampling cell.
Our upsampling cell has four inputs :
$I_{k-1}$, $I_{k-2}$,$I\_{pred}_{k-1}$ and $I_{skip}$.
The inputs $I_{k-1}$, $I_{k-2}$ are the outputs of the last two preceding cells,
$I\_{pred}_{k-1}$ represents a prediction from the previous decoding stage and $I_{skip}$ represents a feature map in the encoder obtained via skip connection.
The inputs $I_{k-1}$, $I_{k-2}$ are upsampled by transposed convolutions whereas the input  $I\_{pred}_{k-1}$ is upsampled by bilinear interpolation (following \cite{dn3}). A schematic of the upsampling cell is shown in Figure \ref{fig:cell_upsampling}. The intermediate nodes in the upsampling cells process all inputs via mixed operations. To be consistent, we use the same set of operations for a mixed operation as DARTS (see Section \ref{sec:search_space}). The outputs of all intermediate nodes are concatenated to form the output of the cell, which is then processed by a 2D convolution to get an upsampled disparity prediction.

\textbf{Search Network.} Compared to standard DARTS which is trained on CIFAR10~\cite{Krizhevsky09} with $32\!\!\times\!\!32$ images, training datasets for disparity estimation~\cite{dn} consist of images which  are about $500$~times larger in terms of pixel amount. Therefore, to feasibly train a search network on a single GPU (Nvidia GTX1080Ti), we downsample the training images by half. Ground truth disparity values are additionally rescaled by a factor of $0.5$.
The encoder part of the search network begins with a stem cell followed by stacked reduction and normal cells.
 The stem cell consists of two standard convolutional layers with kernel sizes \kernel{7} and \kernel{5} and stride $2$ which further downsample the input.
 Similar to the DispNetC architecture~\cite{dn, dn3}, the stem cell and the first reduction cell compose the Siamese part of the encoder which extract features from the left and right rectified stereo view.
 The extracted features are processed by a correlation layer~\cite{Dosovitskiy15}. The correlation layer performs patch comparison between the two feature maps obtained from the Siamese part of the network. Such explicit feature matching helps in significant error reduction~\cite{dn}. The rest of the encoder is formed by stacking normal and reduction cells in an alternating fashion.
 The decoder consists of stacked upsampling cells with skip connections to the encoder.
 The encoder has a total of 6 cells (normal + reduction) with a final downsampling factor of $32$. The decoder consists of three upsampling cells which output predictions at different spatial resolutions. We pre-define each cell to have three intermediate nodes and initialize the first cell to have $24$ channels. Each reduction cell then increases number of channels by a factor of 2. In the decoder, an upsampling cell reduces the number of channels by half with each upsampling step.
 An illustration of our search network is shown in Figure~\ref{fig:search_net}. For training the search network, we optimize the end-point-error (EPE)~\cite{Dosovitskiy15,dn} between the predicted and ground truth disparity maps. A loss term is added for each prediction after an upsampling step. The losses are optimized using the first-order approximation of DARTS as described in Section \ref{sec:search_opt}. We refer to our search network as Dense-DARTS.

\section{Architectures}
After training Dense-DARTS, we extract a normal, a reduction, and an upsampling cell as described in Section \ref{sec:search_and_extraction}.
A schematic of the extraction process is shown in Figure~\ref{fig:search_net}.
A network needs to be built using the extracted cells before it can be trained for final evaluation. In this section we introduce our baseline architecture and present network variants we consider for evaluation.

\label{sec:autodispnet}

\textbf{Baseline architecture.}
\label{sec:baseline}
For a strong baseline we choose a recent state-of-the-art disparity estimation network, DispNet-CSS~\cite{dn3}, which is an improved version of the original DispNet~\cite{dn} manually optimized by an expert. It consists of a stack of three networks, consisting of one DispNet-C~\cite{dn} and two DispNet-S~\cite{dn}. 

\textbf{Single network.}
To compare the performance of the extracted cells, we first build a single network for comparison with the first network in the DispNet-CSS stack. 
In each network the encoder downsamples the input by a factor of $64$ and the output resolution of the decoder is one-fourth of the input resolution. 
%\hl{I think it's HALF resolution? (@Tonmoy it goes up to level 1)}.
For a fair comparison, we use seven encoder cells and four decoder cells to get the same resolutions at the bottleneck and the final layer.
This network is constructed in the same fashion as the search network, as described in Section~\ref{sec:dense_darts}, but with the extracted cells. 
The number of channels for the first cell ($C_{init})$ is set to 42, to match the number of parameters in DispNet-C. We call this network AutoDispNet-C. 

\textbf{Refinement with stacks.}
Using the same configurations as AutoDispNet-C, we construct the AutoDispNet-S architecture by replacing the Siamese part and correlation layer with a single stream of cells. 
In a stacked setting, the AutoDispNet-S network refines a disparity map from the previous network. 
Similar to~\cite{dn3}, the input to the refinement network is a concatenation of warped image, previous disparity prediction and the rectified image pair. The first network provides an initial disparity estimate. Each following network in the stack refines the previous network's output by predicting the residual disparity. The residual disparity is added to the previous network's output to obtain a refined estimate. We denote a stacked AutoDispNet-C and two AutoDispNet-S as AutoDispNet-CSS. The full network stack is shown in Figure~\ref{fig:autodispnet}. 

\textbf{Smaller networks.}
We also experiment with different values of $C_{init}$ to obtain AutoDispNet architectures with different numbers of parameters.
We choose a smaller variant with $C_{init}=18$ for comparison with our baseline. This configuration is denoted as AutoDispNet-css with a lowercase "c" and "s".

\textbf{BOHB variants.}
We also use BOHB to tune the learning rate and weight decay of AutoDispNet architectures. We denote networks of this category by AutoDispNet-BOHB-$(*)$, where $(*)$ stands for C, CS or CSS. Details about training settings are mention in Section \ref{sec:train_search}. 

%\textbf{Hyperparameter search.}

\section{Experiments}
%In this section we describe our experiments showing improved results on disparity estimation with a network obtained by applying Dense-DARTS described in Section \ref{sec:dense_pred}. Additionally, we highlight additional gains in accuracy by optimizing hyperparameters with BOHB~\cite{Falkner18}.

%%%%%%%%%%%%%%%%%%%%%%%%%%%%%%%%%%%%%%%%%%%%%%%%%%%%%%%%%%%%%%%%
\begin{figure*}
\begin{center}
\subcaptionbox{Normal}{\includegraphics[width=0.32\linewidth]{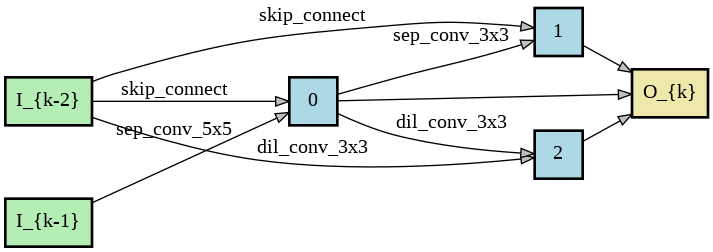}}%
\hfill % <-- Seperation
\subcaptionbox{Reduction}{\includegraphics[width=0.32\linewidth]{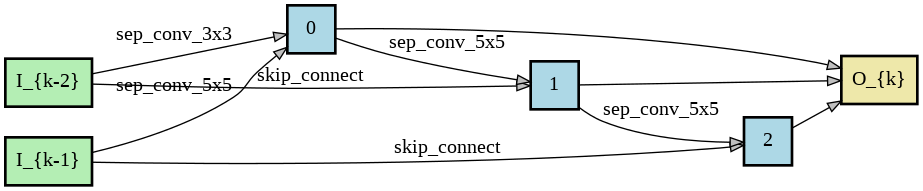}}%
\hfill % <-- Seperation
\subcaptionbox{Upsampling}{\includegraphics[width=0.32\linewidth]{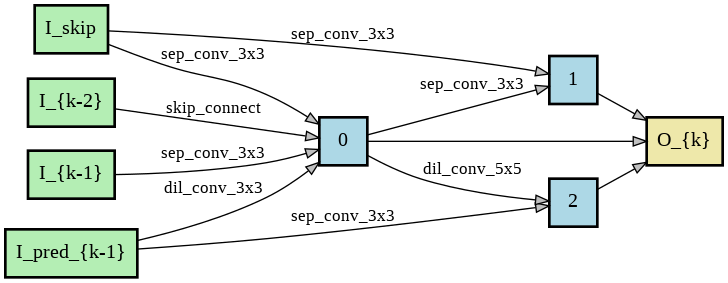}}%
\end{center}
\caption{Learned cells using Dense-DARTS. We visualize the extracted DAG for each cell type. $I_{k-1}$ and $I_{k-2}$ denote the two input nodes and $O_k$ is the output node. The numbered blue nodes depict intermediate nodes, where numbers indicate the depth at which the node was placed in the DAG before extraction.} 
\label{fig:cells}
\end{figure*}

\subsection{Experimental setting}
\label{sec:train_search}

\textbf{Datasets.}
For training our search network we use the standard FlyingThings3D~\cite{dn} dataset which provides ground truth disparity maps generated using rendered scenes.
The dataset consists of train and test splits with $21,818$ and $4,248$ samples respectively. Each sample has a spatial resolution of $960\!\!\times\!\!540$.

\textbf{Training Dense-DARTS.} 
Following ~\cite{darts}, we divide the train split of FlyingThings into two halves. The first and second halves are used to optimize the train and validation losses respectively.
The test split is left untouched to evaluate the extracted architectures at a later stage.
We use the same data augmentation settings as commonly used for training DispNet~\cite{dn,dn3}.
The search network is trained by minimizing the end point error as described in Section~\ref{sec:dense_darts}. 
The train loss is optimized using SGD with base learning rate of $0.025$ and annealing to $0.001$ using the cosine schedule~\cite{loshchilov-ICLR17SGDR}. To optimize the validation loss, we use the Adam optimizer~\cite{Kingma15} with a base learning rate of $1e-4$. 
We add $L2$ regularization on the weight parameters $w$ and architecture parameters $\alpha$ with factors of $3e-4$ and $1e-3$ respectively. Before optimizing $w$ and $\alpha$ alternatingly, we warm start the search network by optimizing only $w$ for $100$k iterations. After the warm-start phase we optimize both $w$ and $\alpha$ for $200$k iterations. We also found that annealing the softmax temperature for the mixed operation leads to slightly better results.

The extracted cells after training the search network are shown in Figure \ref{fig:cells}.
Note that the search process discards all pooling operations. We also see that normal and upsampling cells (which process feature maps at the same or higher spatial resolution) include dilated convolutions, whereas the reduction cell (which downsamples feature maps) consists only of separable convolutions and some skip connections. This observation is in agreement with common usage patterns of operations for dense prediction. For instance, state-of-the art disparity estimation methods~\cite{iresnet,dn3,pwc} are fully convolutional and do not contain any pooling operations. Dilated convolutions have been extensively used to obtain state-of-the art results for semantic segmentation~\cite{Chen17,ChenL18}.

\textbf{Training AutoDispNet architectures.}
For training the AutoDispNet-CSS stack we follow the same training procedure as our baseline architecture~\cite{dn3}. 
For training each refinement network, all previous network weights are frozen~\cite{dn3}. 
Each network is trained for $600k$ iterations using the Adam~\cite{Kingma15} optimizer with a base learning rate of $1e-4$. The learning rate is dropped at $300k$, $400k$, $500k$ with a factor of $0.5$.

\begin{figure}
\begin{center}
\includegraphics[width=\linewidth]{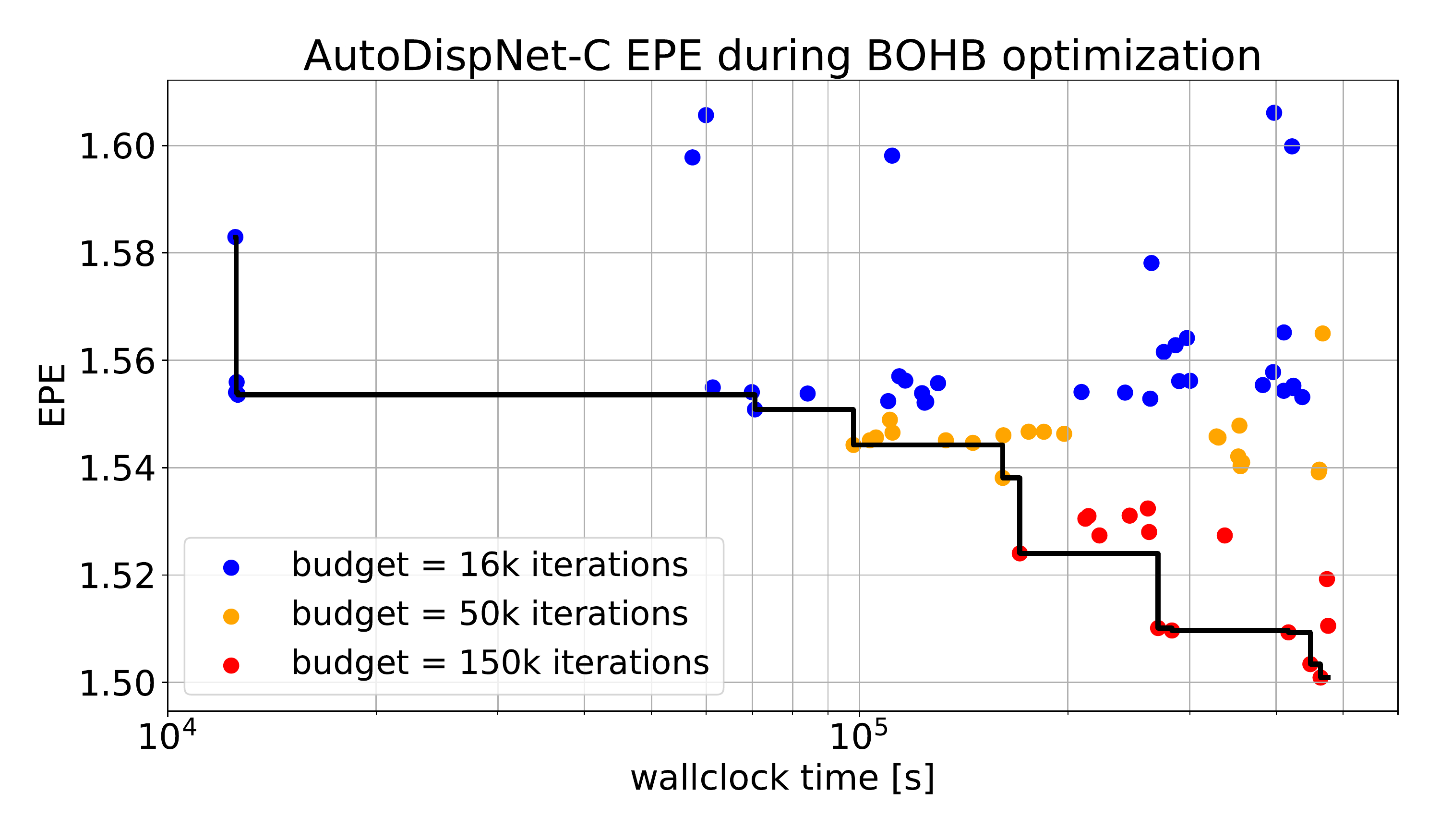}%
\end{center}
\caption{Hyperparameter optimization procedure. AutoDispNet-C EPE on FlyingThings3D of all sampled configurations on the different budgets throughout the BOHB optimization procedure. The black line shows the best performing configurations (incumbent) as a function of time. Note that the value on the x axis is the time when each evaluation finished relative to the BOHB start time, and not the training time per network.}
\label{fig:bohb_runs}
\end{figure}

\textbf{Hyperparameter tuning with BOHB.}
For AutoDispNet-C we optimize the learning rate and weight decay coefficient.
Each function evaluation in BOHB involves training a network with hyperparameters sampled from a configuration space and evaluating it on a validation set.
In this case, we use the test split of FlyingThings3D for validation and use Sintel as a test set.
For small classification networks this usually works because training takes only a few hours. 
However, in our case training is expensive. Training a single network in the stack takes around $3.5$ days on a single Nvidia GTX1080Ti GPU. 
Therefore, to make our function evaluations cheaper, we optimize the learning rate and weight decay for a restart schedule~\cite{loshchilov-ICLR17SGDR}. Specifically, we take a snapshot of the network after $450k$ iterations and restart training with a learning rate sampled by BOHB. The learning rate is annealed to zero at $16.67k$, $50k$ and $150k$ iterations (depending on which of these budgets BOHB evaluates the sampled configurations) following a cosine function \cite{loshchilov-ICLR17SGDR}. This reduces the training cost by a factor of four. The optimized hyperparameter are then used to restart training for successive networks in the stack. We found that using BOHB to tune hyperparameters for the refinement network did not boost performance (we include experimental results in the supplemental).

 We ran BOHB in parallel on 5 GPU workers for a total number of 11 SuccessiveHalving iterations. We used the default BOHB settings with $\eta=3$ and budgets 16.67k, 50k and 150k mini-batch iterations. This is equivalent to 26 full function evaluations on 1 worker, i.e a total of 33.42 GPU days. \figref{fig:bohb_runs} shows the EPE of all sampled configurations throughout the optimization procedure. As we can see, for the budgets of 16.67k and 50k iterations we do not notice any major improvement over time. However, for the maximum number of iterations we observe that BOHB finds a good region in the hyperparameter space and keeps sampling around that area.

\subsection{Results}
\begin{table}[h]
\begin{center}
\resizebox{\linewidth}{!}{  
\begin{tabular}{|l|cc|cc|}
\hline 
\textbf{Architecture}                         & \textbf{FlyingThings3D}      & \textbf{Sintel}  & \textbf{Params} & \textbf{FLOPs} \\
                                              &  (test)             & (train) &  (M)   & (B)\\
\hline 
\hline 
DispNet-C \cite{dn3}                            &  $1.67$     & $3.19$ & $38$       &$75$            \\
\hline 
%Random                              &  $2.02$     & $3.61$ & $12$       &$38$             \\
%AutoDispNet-c                       &  $1.64$     & $2.99$              & $21$       &$48$            \\
AutoDispNet-c                        &  $1.98$     & $3.53$     & $7$        &$16$            \\
AutoDispNet-C                        &  $1.53$     & $2.85$     & $37$       &$61$            \\
AutoDispNet-BOHB-C                   &  $(\textbf{1.51})$     & $\textbf{2.66}$     & $37$       &$61$            \\
\hline 
\end{tabular}
}
\end{center}
\caption {Performance of a single network. We demonstrate improved accuracy of our AutoDispNetC architecture over our baseline DispNetC. End-point errors are shown on the FlyingThings3D and Sintel datasets. The best performance is obtained by optimizing the hyperparameters with BOHB. The parentheses indicate that FlyingThings3D test split is used to optimize hyperparameters.
}
\label{tab:dispnet_single}
\end{table}

\textbf{Single network results.} 
Table~\ref{tab:dispnet_single} shows the result of the automatically optimized DispNet relative to the baseline. AutoDispNet yields significantly stronger numbers with about the same number of parameters. Additional hyperparameter optimization with BOHB yields another improvement on the Sintel dataset~\cite{Butler12}. It is worth noting that the networks were only trained and optimized on the FlyingThings3D dataset, but not on any part of Sintel. This shows that the automated optimization not only overfits better to a particular dataset but improves the general capability of the network.

\begin{table}
 \begin{center}
  \resizebox{0.65\columnwidth}{!}{%
  \begin{tabular}{|l||c|c|c|}%
    \hline
                          & \multicolumn{3}{c|}{\textbf{Number of Networks}} \\
    \cline{2-4}                      
                          & 1  & 2 & 3  \\
    \hhline{|=#=|=|=|}
    DispNet               & C         & CS        &  CSS    \\ 
    EPE                   & $3.19$    & $2.49$    &  $2.36$ \\
    Params                & $38$      & $77$      &  $116$      \\
    FLOPS                 & $75$      & $135$        &  $195$  \\
    \hhline{|=#=|=|=|}
    AutoDispNet           & c         & cs       &  css     \\ 
    %EPE                   & $2.99$    & $2.35$   &  ?     \\
    EPE                   & $3.53$    & $2.80$   &  $2.54$     \\
    Params                & $7$       & $14$     &  $21$             \\
    FLOPS                 & $16$      & $30$     &  $44$     \\
    \hline 
    \hhline{|=#=|=|=|}
    AutoDispNet     & C               & CS      &  CSS  \\ 
    EPE             &  $2.85$         & $2.30$  &  $2.14$     \\
    Params          & $37$            & $74$    &  $111$     \\
    FLOPS           & $61$            & $110$   &  $160$     \\
    \hline
    \hhline{|=#=|=|=|}
    AutoDispNet-BOHB     & C               & CS             &  CSS  \\ 
    EPE             & $\mathbf{2.66}$ & $\mathbf{2.14}$     &  $\textbf{1.97}$     \\
    Params          & $37$            & $74$    &  $111$     \\
    FLOPS           & $61$            & $110$   &  $160$     \\
    \hline 
  \end{tabular}
  }
  \end{center}

\caption{Performance across the stack. We show improved performance of AutoDispNet architectures across the network stack. End point errors are reported for the Sintel dataset. AutoDispNet-CS matches the baseline performance of three networks with only a single refinement network.
The AutoDispNet-BOHB-CS variant outperforms the three network baseline in the second network itself.
  \label{tab:dispnet-stacks}
 }
\end{table}

\textbf{Stacked network results.} 
For state-of-the-art performance on disparity estimation, it is necessary to stack multiple networks. Table~\ref{tab:dispnet-stacks} shows that the benefits of automated optimization also carry over to the large stacked networks. There is a significant improvement with both the architecture optimization and the hyperparameter optimization. The results also reveal that a stack of two networks is already more accurate than a stack of three networks with the baseline. Also the small version of AutoDispNet-css is competitive with DispNet-CS, but runs with 3 times less FLOPS. %faster. 

\textbf{Comparison to the state of the art.}
Although we considered only a limited set of published architectural choices for AutoDispNet, Table~\ref{tab:compare_disp} reveals that it is competitive with the state of the art on the common public benchmarks. Only PSMNet with its coarse-to-fine strategy performs better on KITTI 2012, but worse on KITTI 2015.

\begin{table}[ht]

\centering 
\newcommand{\rowline} {\hhline{|~||~|~~~~|}} 
\newcommand{\headerline} {\hhline{|=||=|====|}} 
\newcommand{\headertitle}[1]{\textbf{#1} &&&&&\\}

\resizebox{\linewidth}{!}{%
\begin{tabular}{|l||c|cccc|}
% --- begin header
\hhline{|-||-|----|}
\textbf{Method} 
&\textbf{Sintel}
&\multicolumn{2}{c}{\textbf{KITTI}}
&\multicolumn{2}{c|}{\textbf{KITTI}}\\
\hhline{|~||~|~~~~|}
%empty
& (clean)
&\multicolumn{2}{c}{(2012)}
&\multicolumn{2}{c|}{(2015)}
\\
%empty
& AEE
& AEE & Out-noc
& AEE & D1-all
\\
& \textit{train}  
& \textit{train} & \textcolor{black}{\textit{test}} 
& \textit{train} & \textcolor{black}{\textit{test}}

\\
\headerline 
% --- end header

% ########## Standard approaches ##############
%\textbf{Standard}
%&&&&&&&&&\\
\headertitle{Others}
\headerline 
% --- begin method
SGM ~\cite{sgm}
& $19.62$    % Sintel train
& $10.06$ & -  % KITTI 12 train,test
& $7.21$  & $10.86\%$   % KITTI 15 train,test
\\
%% --- begin method
DispNet-CSS \cite{dn3}
& $2.33$  % Sintel from tensorflow
& $1.40$ & -  % KITTI 12 train,test tf
& $1.37$ & -  % KITTI 15 train,test tf
\\
% --- end method
% --- begin method
DispNet-CSS-ft \cite{dn3}
& $5.53$   % Sintel tf
& $(0.72)$  & $1.82\%$  % KITTI 12 train,test (sc05a) tf
& $(0.71)$  & $2.19\%$  % KITTI 15 train,test (sc05a) tf 
\\
% --- end method 
%% --- begin method
iResNet-i2 ~\cite{iresnet}
& - % Sintel
& -  & $1.71\%$ % KITTI 12 train,test
&  - & - 	  % KITTI 15 train,test
\\
%--- begin method
EdgeStereo\cite{edge_stereo}
& -  % Sintel
& -  & -  % KITTI 12 train,test
& -  & $\mathbf{2.16\%}$	  % KITTI 15 train,test
\\
%% --- begin method
PSMNet	~\cite{psmnet}
& -  % Sintel
& -  & $\mathbf{1.49 \%}$  % KITTI 12 train,test
& -  & $2.32\%$  % KITTI 15 train,test
\\
% --- end method 
% --- end method 
%% --- begin method
GC-Net \cite{gcnet}
& -               % Sintel
& -  & $1.77\%$  % KITTI 12 train,test
& -  & $2.87\%$  % KITTI 15 train,test
\\
% --- end method 
%% --- begin method
SegStereo ~\cite{SegStereo}
& -  % Sintel
& -   & $1.68\%$  % KITTI 12 train,test
& -   & $2.25\%$  % KITTI 15 train,test
\\
% --- end method 

% ########## END CNN  approaches ##############

\headerline 
% ########## BEGIN Joint Estimation ##############
\headertitle{Ours}
\headerline 
% --- begin method
%s12c1
AutoDispNet-css 
& $2.53$    % Sintel 
& $1.03$ & -  % KITTI 12 train,test tf
& $1.19$ & -  % KITTI 15 train,test tf
\\
% --- begin method
%s12c1
AutoDispNet-CSS 
& $2.14$    % Sintel 
& $0.93$ & -  % KITTI 12 train,test tf
& $1.14$ & -  % KITTI 15 train,test tf
\\

% --- begin method
%s13a
AutoDispNet-BOHB-CSS
& $\textbf{1.97}$  % Sintel from tensorflow
& $0.94$ & -  % KITTI 12 train,test tf
& $1.15$ & -  % KITTI 15 train,test tf
\\
% --- end method
% --- begin method
AutoDispNet-BOHB-CSS-ft
& $10.55$   % Sintel tf
& $(0.45)$       & $1.70\%$  
& $(0.50)$       &  $2.18\%$  
\\
% --- end method

% --- end method

% ########## END Standard approaches ##############

\hhline{|-||-|----|}

\end{tabular}

}

\caption{
\label{tab:compare_disp}
Benchmark results. We compare performance of our networks on Sintel and KITTI datasets. For Sintel and KITTI train sets, we report the average end-point error (AEE). 
Out-noc and D1-all are metrics used to rank methods on the KITTI'12 and KITTI'15 leader boards. Out-noc is the percentage of outliers exceeding an error threshold of 3px. D1-all is the same metric but applied on all regions (occ and non-occ). Entries enclosed by parentheses indicate if they were finetuned for the evaluated dataset. On KITTI'15 we are comparable to our baseline. On KITTI'12 we outperform the baseline with a significant margin.
} 
\end{table}

% KITTI 12
%\begin{tabular}{c | c | c | c}
%{\bf Error} & {\bf Out-Noc} & {\bf Out-All} & {\bf Avg-Noc} & {\bf Avg-All}\\ \hline
%2 pixels & 2.54 \% & 3.05 \% & 0.5 px & 0.5 px\\
%3 pixels & 1.70 \% & 2.05 \% & 0.5 px & 0.5 px\\
%4 pixels & 1.31 \% & 1.58 \% & 0.5 px & 0.5 px\\
%5 pixels & 1.06 \% & 1.28 \% & 0.5 px & 0.5 px
%\end{tabular}

%KITTI 15
%\begin{tabular}{c | c | c}
%{\bf Error} & {\bf D1-bg} & {\bf D1-fg} & {\bf D1-all}\\ \hline
%All / All & 1.94 \% & 3.37 \% & 2.18 \%\\
%All / Est & 1.94 \% & 3.37 \% & 2.18 \%\\
%Noc / All & 1.80 \% & 2.98 \% & 2.00 \%\\
%Noc / Est & 1.80 \% & 2.98 \% & 2.00 \%
%\end{tabular}

\subsection{Applicability to other tasks}
\begin{table}[ht]
\begin{center}
\resizebox{\columnwidth}{!}{%
\begin{tabular}{|l|c|c|c|c|c|}
\hline 
\textbf{Method}            & \textbf{Params} & \textbf{Abs. rel} & \textbf{Sqr. rel} & \textbf{Rmse}  & \textbf{\begin{tabular}[c]{@{}c@{}}Rmse (log)\end{tabular}} \\\hline
\multicolumn{6}{|l|}{\textit{\textbf{SUN3D}}}\\\hline
    %&                 &                   &                   &                & \\\hline
Laina \etal~\cite{laina} & 63M             & 0.272             & 0.248             & 0.703          & 0.500                                                         \\
AutoDepth-S               & 63M             & \textbf{0.234}    & \textbf{0.202}    & \textbf{0.602} & \textbf{0.453}                                                \\
AutoDepth-s               & 38M             & 0.234             & 0.210             & 0.614          & 0.518                                                         \\\hline
%\textit{\textbf{NYU-v2-test}} &                 &                   &                   &                &      \\\hline
\multicolumn{6}{|l|}{\textit{\textbf{NYU-Depth-V2}}}\\\hline
Laina \etal~\cite{laina} & 63M             & \textbf{0.127}    & \textbf{-}        & \textbf{0.573} & \textbf{0.195}                                                \\
AutoDepth-BOHB-S           & 63M             & 0.170             & 0.141             & 0.599          & 0.216   \\\hline 
\end{tabular}
}
\end{center}

\caption{
\label{tab:depth}
Results on single view depth estimation. Auto-Depth represents a network found using Dense-DARTS.  (For details about the metrics see \cite{eigen}) 
}
\end{table}

We also tested our approach on single view depth estimation, another dense prediction task and compare with Laina \etal~\cite{laina}, a state-of-the art single view depth estimation method. The results are shown in Table \ref{tab:depth}. On SUN3D we obtain an improvement over the baseline, however the results on NYU dataset are slightly worse. For more details please see the supplement.

\section{Conclusion}
AutoDispNet extends efficient neural architecture search to large-scale dense prediction tasks, in particular U-Net-like architectures.
It also leverages hyperparameter tuning by running BOHB on the selected architecture. Results show that this sort of optimization leads to substantial improvements over a manually optimized baseline and reaches state-of-the-art performance on the well-optimized task of disparity estimation.
This optimization did not require a huge compute center but was run on common compute hardware, i.e., it can be run by everybody.
The total time taken to obtain the AutoDispNet-BOHB-CSS architecture is approximately $42$ GPU days.

\section*{Acknowledgements}

This study was supported by the German Federal Ministry of Education and
Research via the project DeToL and by the German Research Foundation
under Germany's Excellence Strategy (CIBSS-EXC-2189).

{\small
\bibliographystyle{ieee_fullname}
\bibliography{ref}
}

\newcommand{\beginsupplement}{%
        \setcounter{table}{0}
        \renewcommand{\thetable}{S\arabic{table}}%
        \setcounter{figure}{0}
        \renewcommand{\thefigure}{S\arabic{figure}}%
        \setcounter{section}{0}
     }
     
\newpage 
\clearpage

\beginsupplement 
\twocolumn[{\centering{ \Huge Supplementary Material}\vspace{3ex}
%\smallbreak
%\blindtext
%\medbreak
\vspace{2ex}
}]

\section{Learning curves}
\begin{figure}[t]
\begin{center}
\includegraphics[width=\linewidth]{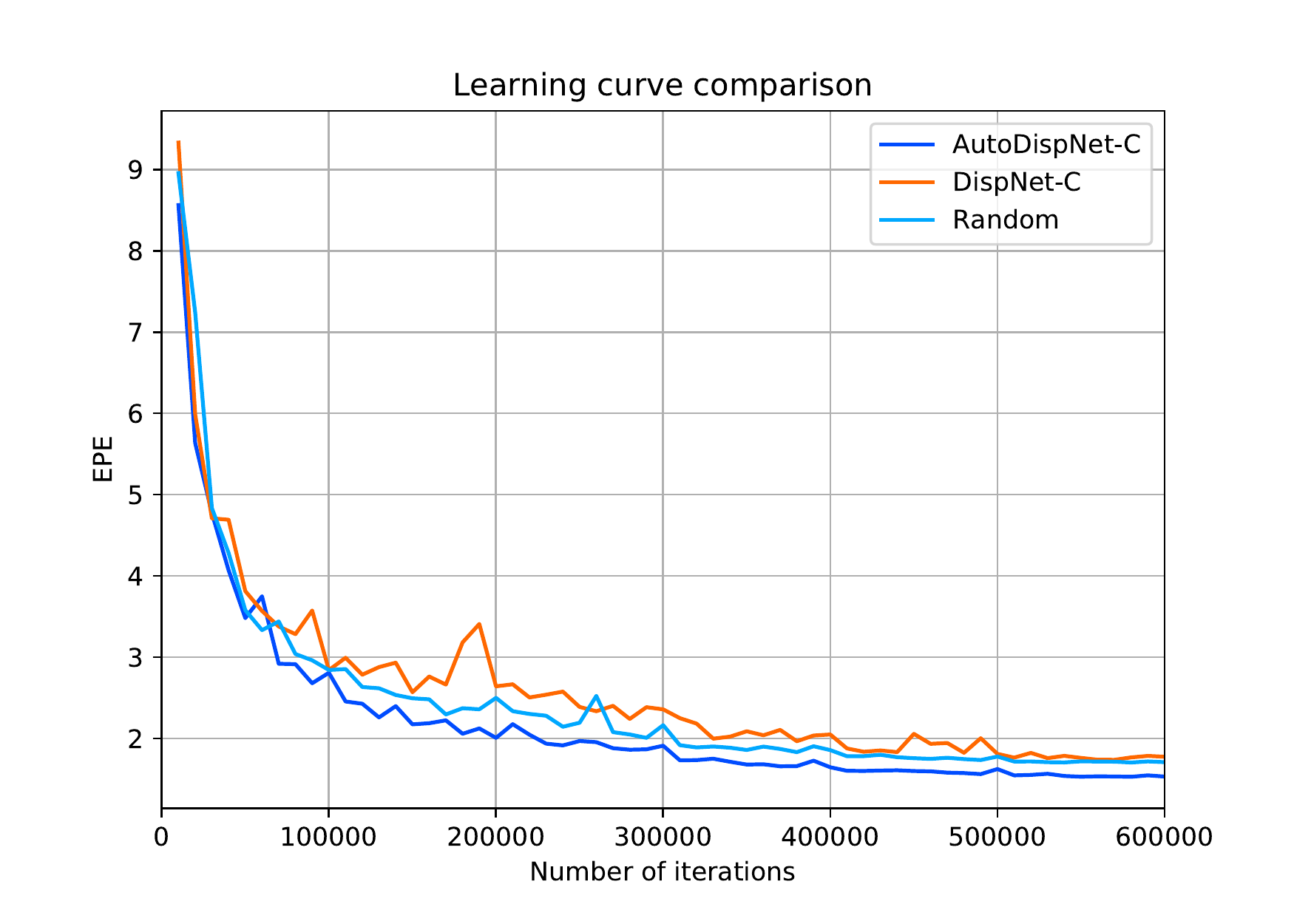}%
\end{center}
\caption{Learning curve comparison. We compare the learning curve of AutoDispNet-C with the baseline DispNet-C and an architecture built with random cells sampled from our search space (denoted by Random). The evolution of EPE over number of iterations is shown for the FlyingThings3D dataset (test split). }
\label{fig:lcurves}
\end{figure}

Figure \ref{fig:lcurves} shows the learning curve (evolution of EPE over number of iterations) of AutoDispNet-C.
In addition to the baseline (DispNet-C), we also compare the learning curve of a random cell architecture. We randomly sample cells from the search space and stack them in the same fashion as AutoDispNet-C (see section \ref{sec:autodispnet} of main paper). We sample four times and build four different random architectures. After training, we pick the best random architecture based on the validation performance on FlyingThings3D. All networks are trained using the same settings as the baseline. Learning curve of the best random architecture is shown in Figure \ref{fig:lcurves}. We observe that AutoDispNet-C clearly outperforms the random architecture and the baseline. We also observe that the random architecture is comparable to the baseline. Our observation is similar to Liu \etal\cite{darts} on classification, where they also report a surprisingly strong performance for random architectures. 

\section{Performance of smaller networks}
\begin{figure*}[t]
\begin{center}
\subcaptionbox{EPE vs Params \label{fig:epeparams}}{
\includegraphics[width=0.45\linewidth]{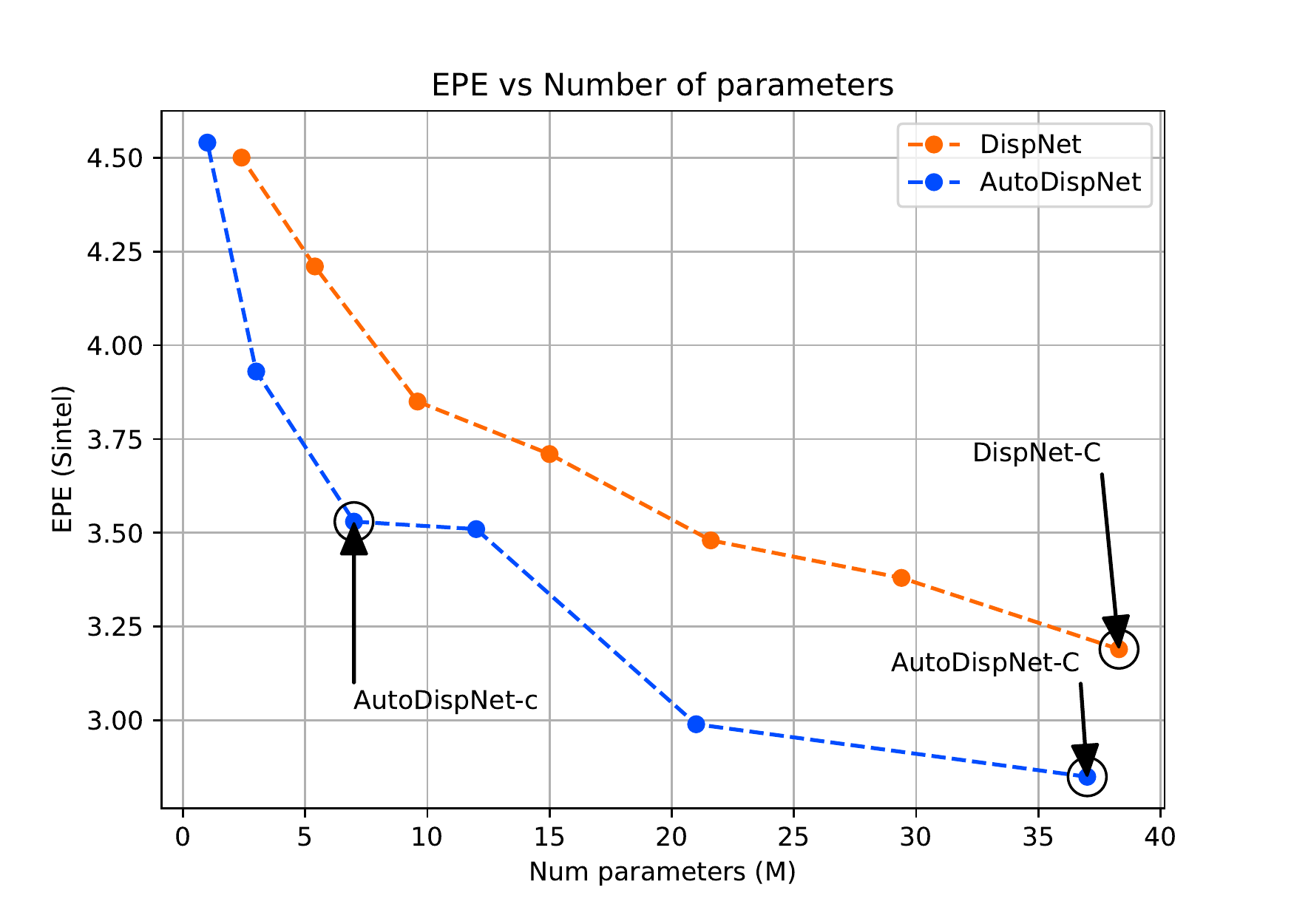}%
} 
%\hfill 
\subcaptionbox{EPE vs FLOPS \label{fig:epeflops}}{
\includegraphics[width=0.45\linewidth]{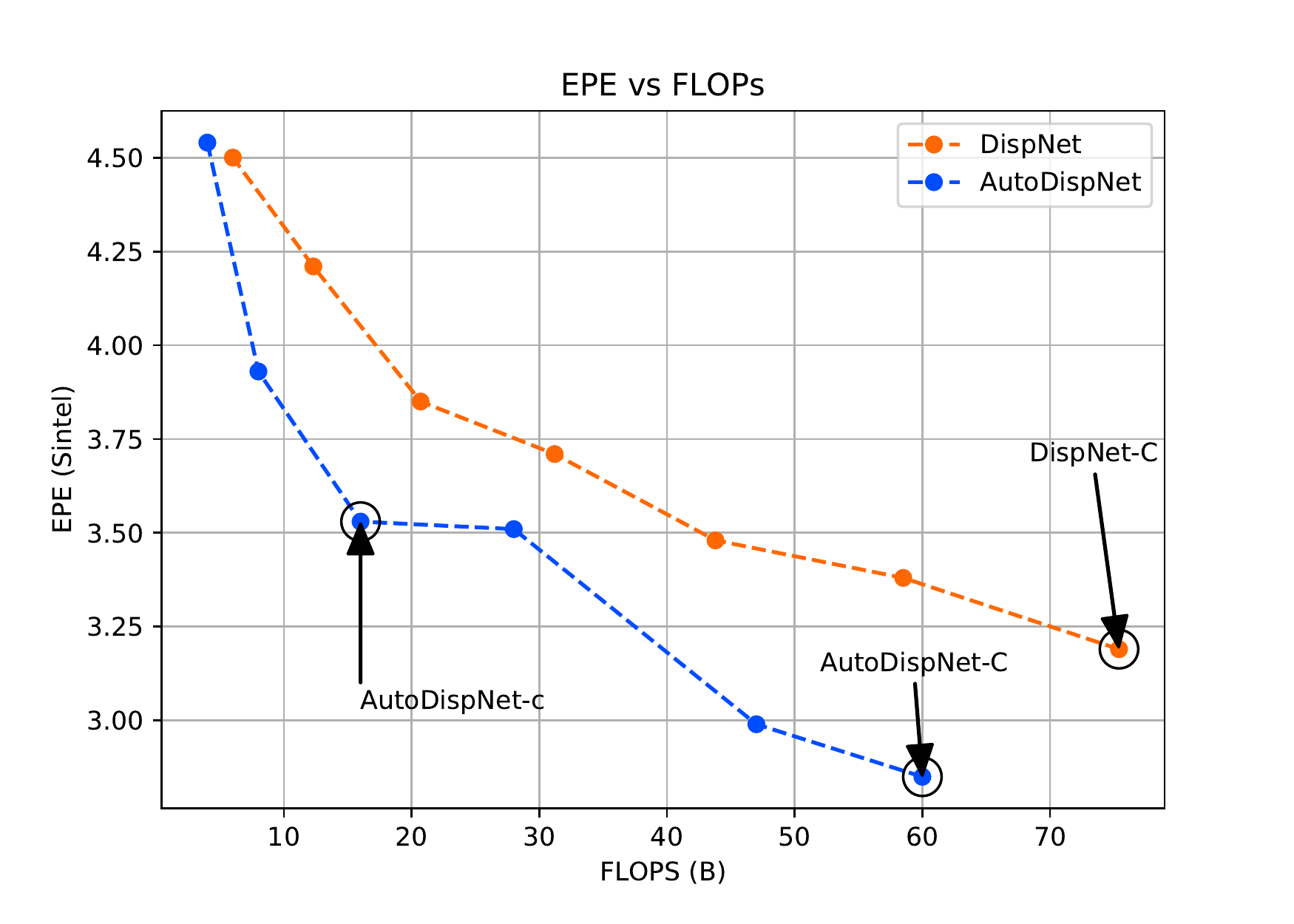}%
} 
\end{center}
\caption{Performance of smaller networks. We compare the test performance of smaller DispNet and AutoDispNet architectures. In Figure \ref{fig:epeparams}, we see that AutoDispNet architectures have a lower error with reduced number of parameters compared to the baseline. A similar trend is observed on comparing the EPE with respect to FLOPS ( Figure \ref{fig:epeflops}).} The EPE is shown for the Sintel dataset. 
\label{fig:perf_compare}
\end{figure*}

We train networks of reduced capacities for both AutoDispNet-C and DispNet-C architectures. 
For DispNet-C smaller networks are obtained by multiplying the number of channels for each layer by fixed factor (similar to~\cite{Ilg17}). 
Smaller variants of AutoDispNet-C are obtained by reducing the number of channels ($C_{init}$) for the first cell. A comparison of EPE vs number of parameters and EPE vs FLOPS is shown in Figure \ref{fig:perf_compare}. 

\section{Optimizing the refinement network}

\begin{table}[h]
\begin{center}
\resizebox{0.9\linewidth}{!}{  
\begin{tabular}{|c|c|c| }
\hline 
\multicolumn{2}{|c|}{\textbf{Network stack}} & \textbf{EPE} \\
\cline{0-1} 
 C                          & S            & (Sintel)  \\
\hline
Dense-DARTS                 &   reuse cells                     &  $2.30$    \\
Dense-DARTS                 &   Dense-DARTS                     &  $2.32$    \\
\hline 
Dense-DARTS $+$ BOHB            &   reuse cells $+$ hyperparams       &  $2.14$    \\
Dense-DARTS $+$ BOHB            &   reuse cells $+$ BOHB              &  $2.16$    \\
\hline 
\end{tabular}
}
\end{center}
\caption {We show the results of optimizing cells and hyperparameters of the refinement network in a stack containing two networks (AutoDispNet C and S).
First row shows a network where cells for the first network are learned using Dense-DARTS and the refinement network reuses these cells. In the second row, we learn new cell structures of the refinement network using Dense-DARTS. In the third row, we learn cells for the first network and tune the hyperparameters using BOHB. In this case, the refinement network reuses both cells and hyperparameters. In the fourth row, we learn new hyperparameters for the refinement network using BOHB but still use the same cell structures as the first network.}
\label{tab:refine_net}
\end{table}

In a stacked setting, the refinement network predicts the residual for correcting errors in predictions from the previous network.
Since this task is different from predicting disparity from scratch, we trained a search network to learn specialized cells for the refinement task. 
However, we found that learning cells for the refinement network did not improve performance over reusing cells learned for the first network. The same argument can also be made for optimizing hyperparameters of the refinement network using BOHB. Surprisingly, even BOHB did not yield improvements over reusing hyperparameters learned for the first network. We show our experimental results in Table \ref{tab:refine_net}. We conjecture that the refinement task is much simpler than estimating disparity from scratch and optimizing cells or hyperparameters is trivial in this case.

\section{Finetuning on the KITTI dataset}
\begin{figure}[ht]
\begin{center}
\includegraphics[width=\linewidth]{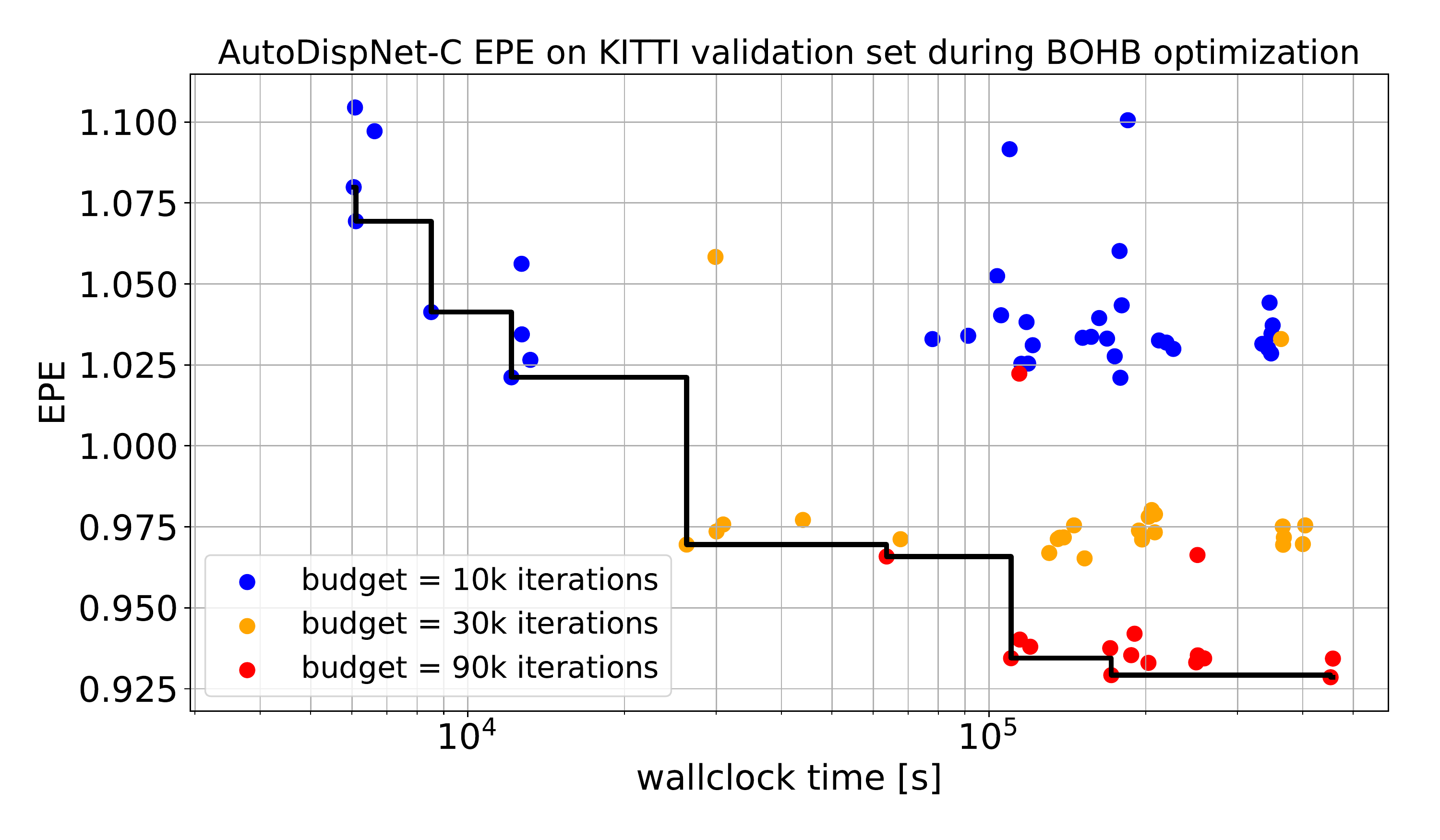}%
\end{center}
\caption{Hyperparameter optimization on KITTI. AutoDispNet-C EPE of all sampled configurations on the different budgets throughout the BOHB optimization procedure. The black line shows the best performing configurations (incumbent) as a function of time.}
\label{fig:bohb_runs_kitti}
\end{figure}

For finetuning on KITTI, we optimize the learning rate and weight decay coefficient using BOHB for the first network in the stack.
For running BOHB, we take all samples from KITTI'12 and KITTI'15 datasets and use $70\%$ of the mixture for training. The remaining $30\%$ of the samples are used for validation. We ran BOHB in parallel on 5 GPU workers for a total number of $10$ SuccessiveHalving iterations. We used the default BOHB settings with $\eta=3$ and budgets $10k$, $30k$ and $90k$ mini-batch iterations. For each budget the learning rate is annealed to zero using a cosine schedule. \figref{fig:bohb_runs_kitti} shows the EPE of all sampled configurations throughout the optimization procedure. 
The optimized hyperparameters are then used to finetune the successive networks in the stack. For the last network, we add two more decoding stages to go to full resolution. Here, we use transposed convolutions instead of upsampling cells because applying the cell structure at higher resolutions becomes computationally expensive.

\section{Single view depth estimation}
To evaluate on single view depth estimation, we used the proposed extension of DARTS and compare our results with the competitive method by Laina \etal \cite{laina}, which uses a ResNet based encoder-decoder with hand-designed upsampling blocks. For a fair comparison, we evaluated both architectures by training them on a subset of the SUN3D dataset using the same hyperparameters and loss function. Please note that in this setting, the siamese part of the network is replaced with a single stream. The extracted architecture is then fine-tuned on $\sim\!\!10,000$ samples from the NYU train dataset using BOHB (optimizing the learning rate and weight decay).

\section{More details on BOHB}\label{sec: bohb_sup}
%As briefly described in Section \ref{sec:bohb},
BOHB \cite{Falkner18} combines Bayesian Optimization (BO) and Hyperband (HB) \cite{li17} in order to exhibit strong anytime and final performance. BOHB follows the same strategy as HB to allocate resources to configurations calling the SuccessiveHalving (SH) \cite{Jamieson16} subroutine repeatedly on its inner loop.  Refer to Algorithm \ref{alg: HB} for a pseudo-code for Hyperband.

\begin{figure}[t]
\begin{center}
\includegraphics[width=\linewidth]{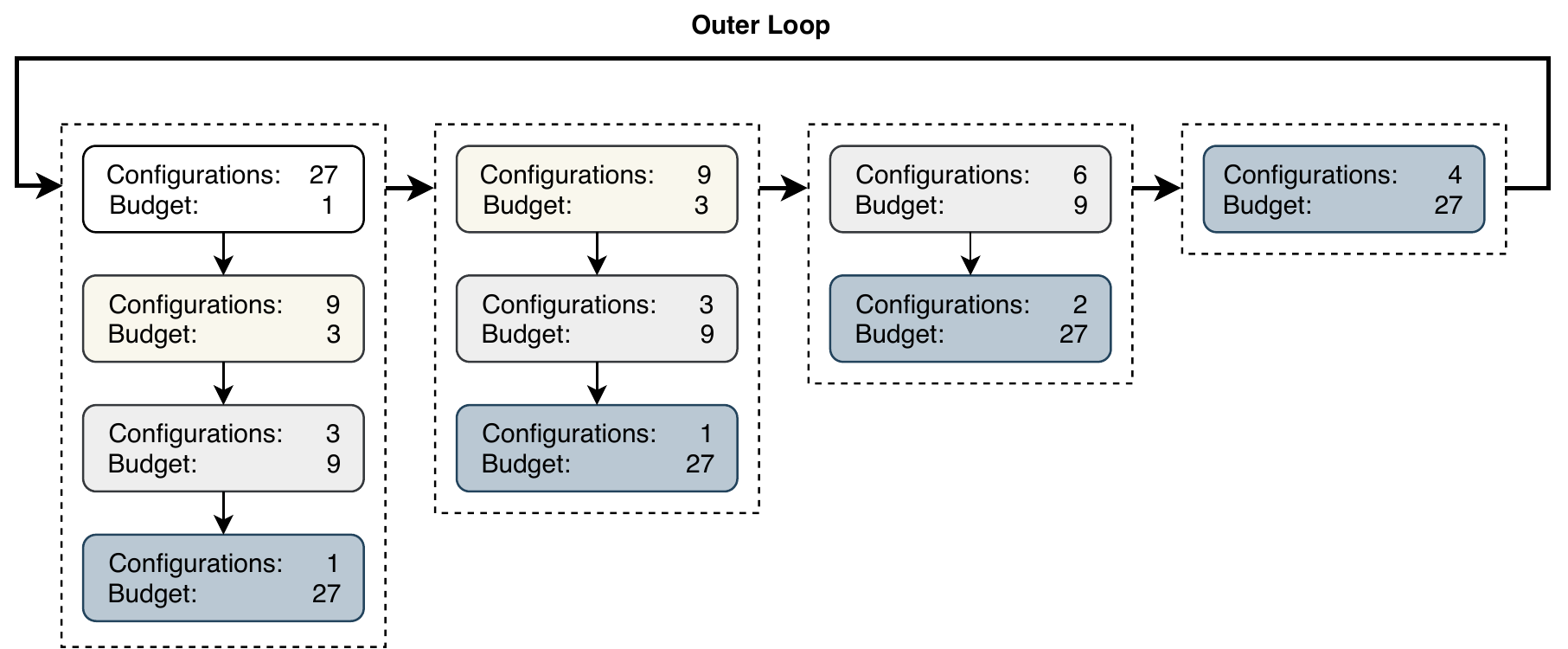}%
\end{center}
\caption{Hyperband inner and outer loops.  Hyperband runs SuccessiveHalving on its inner loop with a initial budget and number of starting configurations determined on its outer loop such that the total budget in every SuccessiveHalving run is the same.}
\label{fig:bohb_flowchart}
\end{figure}

On the outerloop HB samples uniformly $N$ random configurations from the hyperparameter search space (lines \ref{line3}-\ref{line4}).
Afterwards, SH evaluates these $N$ configurations (line \ref{line7}) on the smallest available budget for this outerloop iteration (line \ref{line5}) and advances the best $1/\eta$ performing configurations (line \ref{line8}) to evaluate on a higher budget (increased by a factor of $\eta$; line \ref{line9}). This process goes on until the maximum available budget is reached (line \ref{line6}). As an example, suppose SH starts with a maximum $N=27$ number of sampled hyperparameter configurations for training a neural network with a minimum budget of $b_{min}=1$ epoch (first SH innerloop in \figref{fig:bohb_flowchart}). With an $\eta = 3$ the next iteration of SH would start the best $N/\eta = 9$ configurations evaluated on some validation set with the second budget $\eta \cdot b_{min} = 3$ epochs. This will continue until only one configuration is evaluated for $b_{max} = 27$ epochs.

\begin{algorithm}[t]
\SetKwInOut{Input}{input}\SetKwInOut{Output}{output}
\SetAlgoLined
\Input{min/max budgets $b_{min}$, $b_{max}$, $\eta$}
 $s_{max} = \lfloor log_{\eta}\frac{b_{max}}{b_{min}} \rfloor$\;
 \footnotesize{\tcp{Begin HB outerloop}}
 \For{$s \in \{ s_{max}, s_{max}-1,...,0 \}$}{
  $N = \lceil \frac{s_{max}+1}{s+1} \cdot \eta^s \rceil$\;\label{line3}
  sample $N$ configurations $C = \{ c_1, c_2,...,c_N \}$\;\label{line4}
  \footnotesize{\tcp{Initial budget for SH}}
  $b = \eta^{-s} \cdot b_{max}$\;\label{line5}
  \footnotesize{\tcp{Start SH innerloop}}
  \While{$b \leq b_{max}$}{\label{line6}
   \footnotesize{\tcp{Evaluate all configurations in $C$ for the given budget}}
   $L = \{ \Tilde{f}(c, b) | c \in C \}$\;\label{line7}
   \footnotesize{\tcp{Keep only the best $\lfloor \lvert C/\eta \rvert \rfloor$ ones}}
   $C = top\_k(C, L, \lfloor \lvert C/\eta \rvert \rfloor)$\;\label{line8}
   \footnotesize{\tcp{Increase budget by a factor of $\eta$}}
   $b = \eta \cdot b$\;\label{line9}
   }
  }
 \caption{Hyperband pseudocode}
 \label{alg: HB}
\end{algorithm}

In order to account for the very aggressive evaluations with many configurations on the smallest budget (as done in the first SH innerloop), HB resets SH to start with a smaller degree of aggressiveness, i.e. evaluating the new sampled configurations on a larger initial budget (lines \ref{line3}-\ref{line5} in Algorithm \ref{alg: HB}; illustrated in the second innerloop of  \figref{fig:bohb_flowchart}). Nevertheless, the number of configurations $N$ sampled in every HB outerloop iteration (line \ref{line3} in Algorithm \ref{alg: HB}) is chosen such that the same total budget is assigned to each SH run.

Even though BOHB relies on HB to balance the number of configurations it evaluates and the resources assigned to each configuration, it replaces the random sampling in line \ref{line3} of Algorithm \ref{alg: HB} by a model-based sampling, where the model is build by the configurations evaluated so far. The strong final performance of BOHB arises from the model-based guided search, which effectively focuses more attention to regions in space where good configurations lie.

\begin{figure*}
\begin{center}
\includegraphics[width=0.33\textwidth]{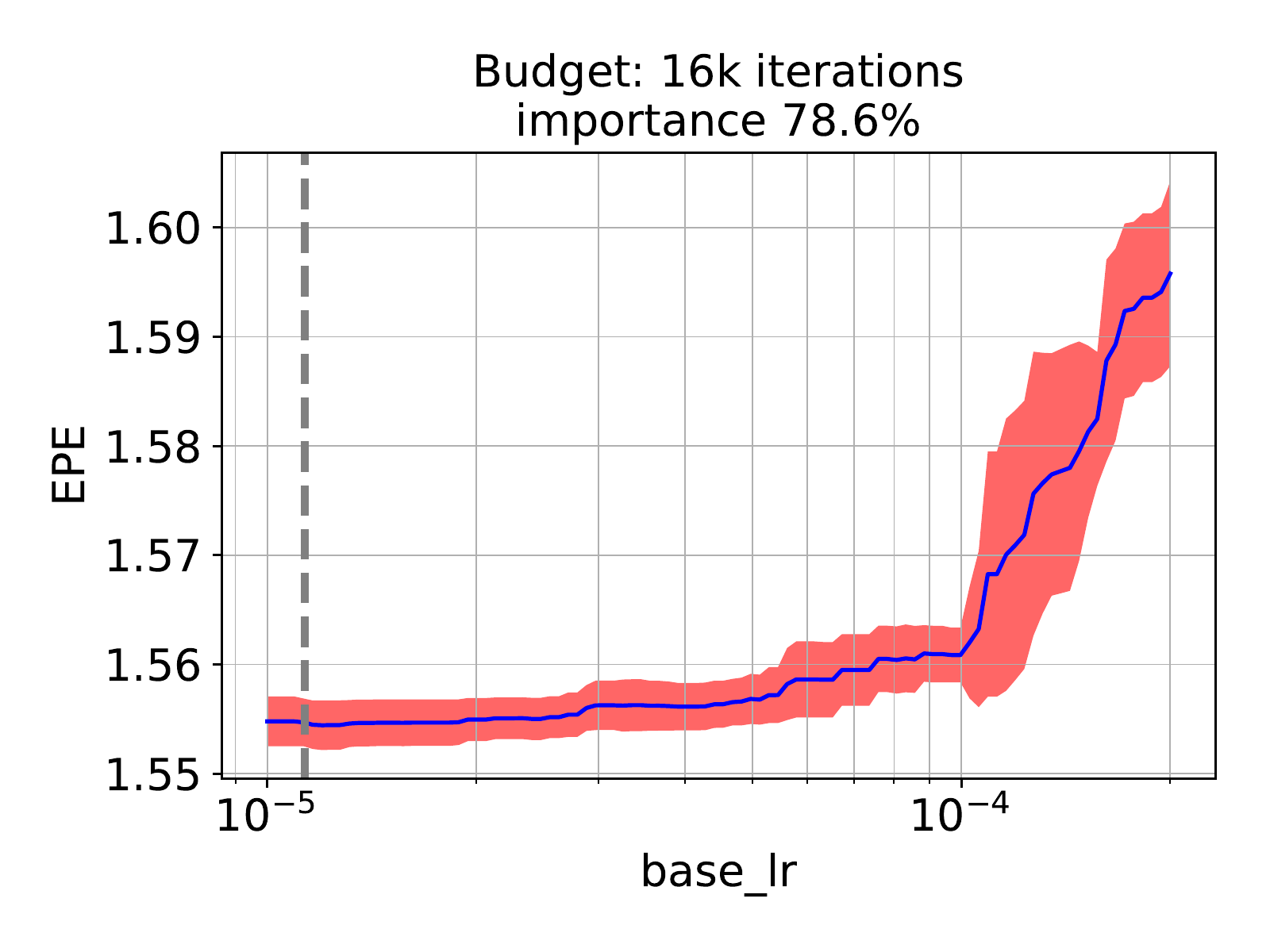}%
%\hfill % <-- Seperation
\includegraphics[width=0.33\textwidth]{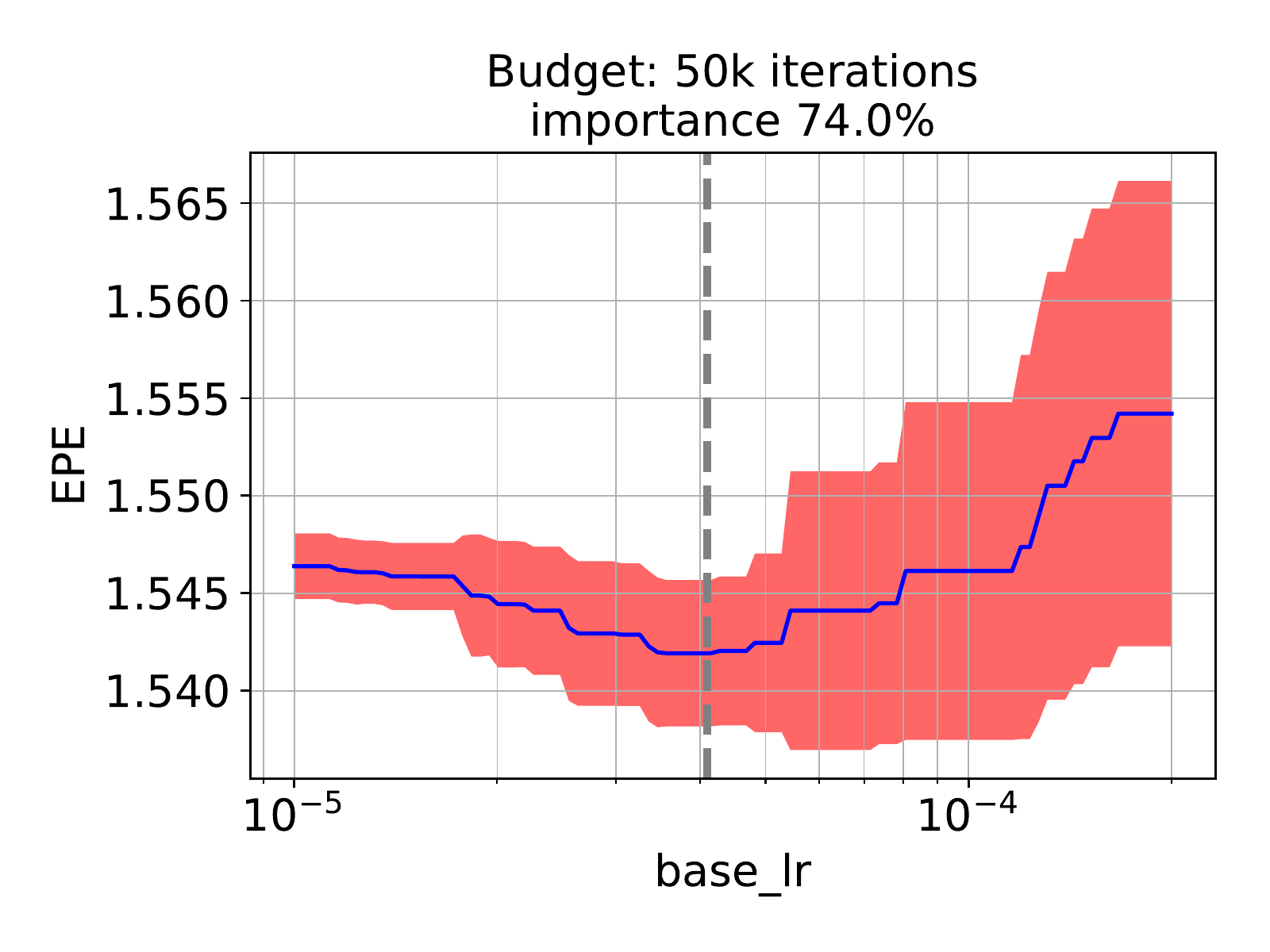}%
%\hfill % <-- Seperation
\includegraphics[width=0.33\textwidth]{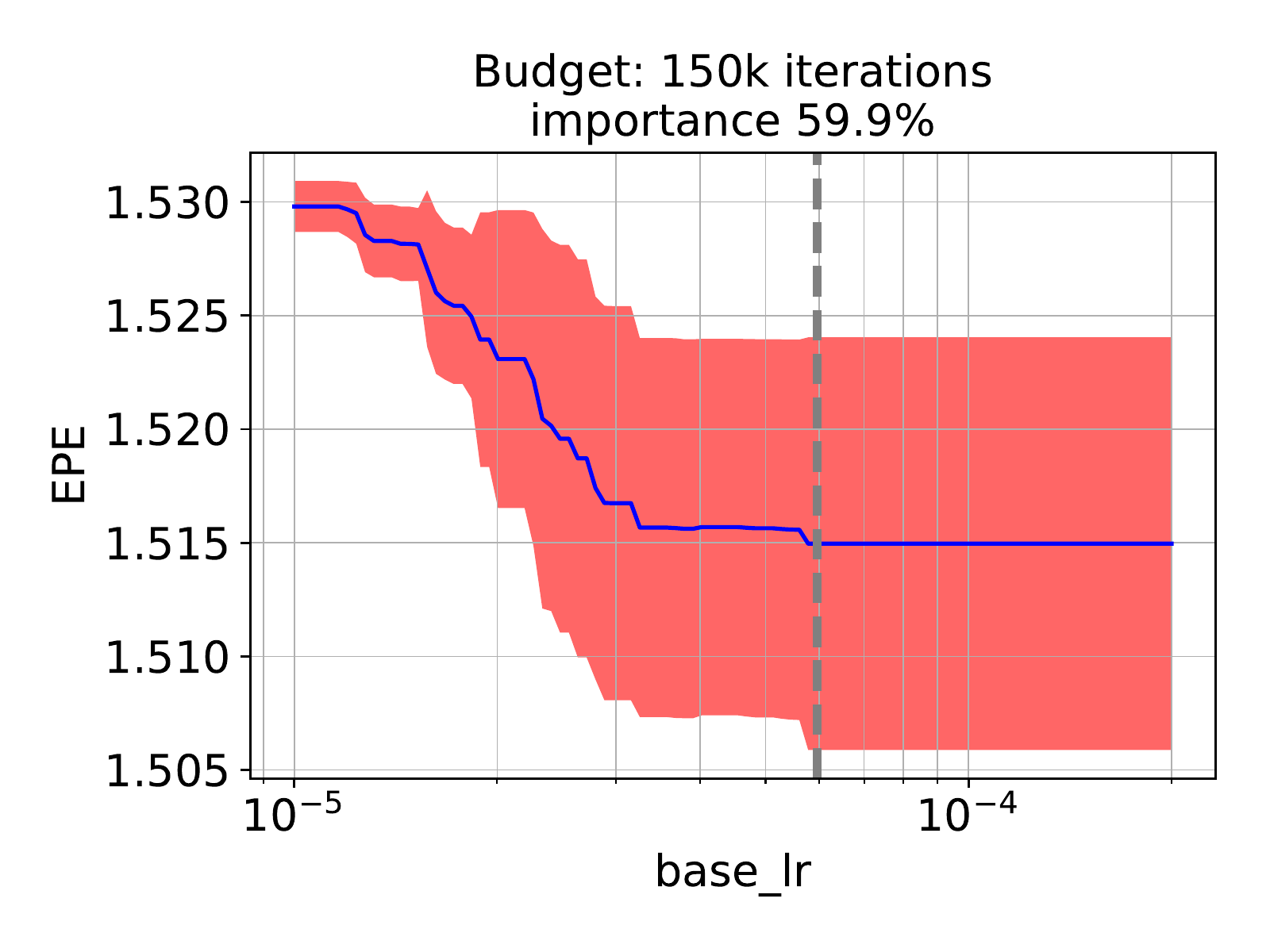}\\%
\includegraphics[width=0.33\textwidth]{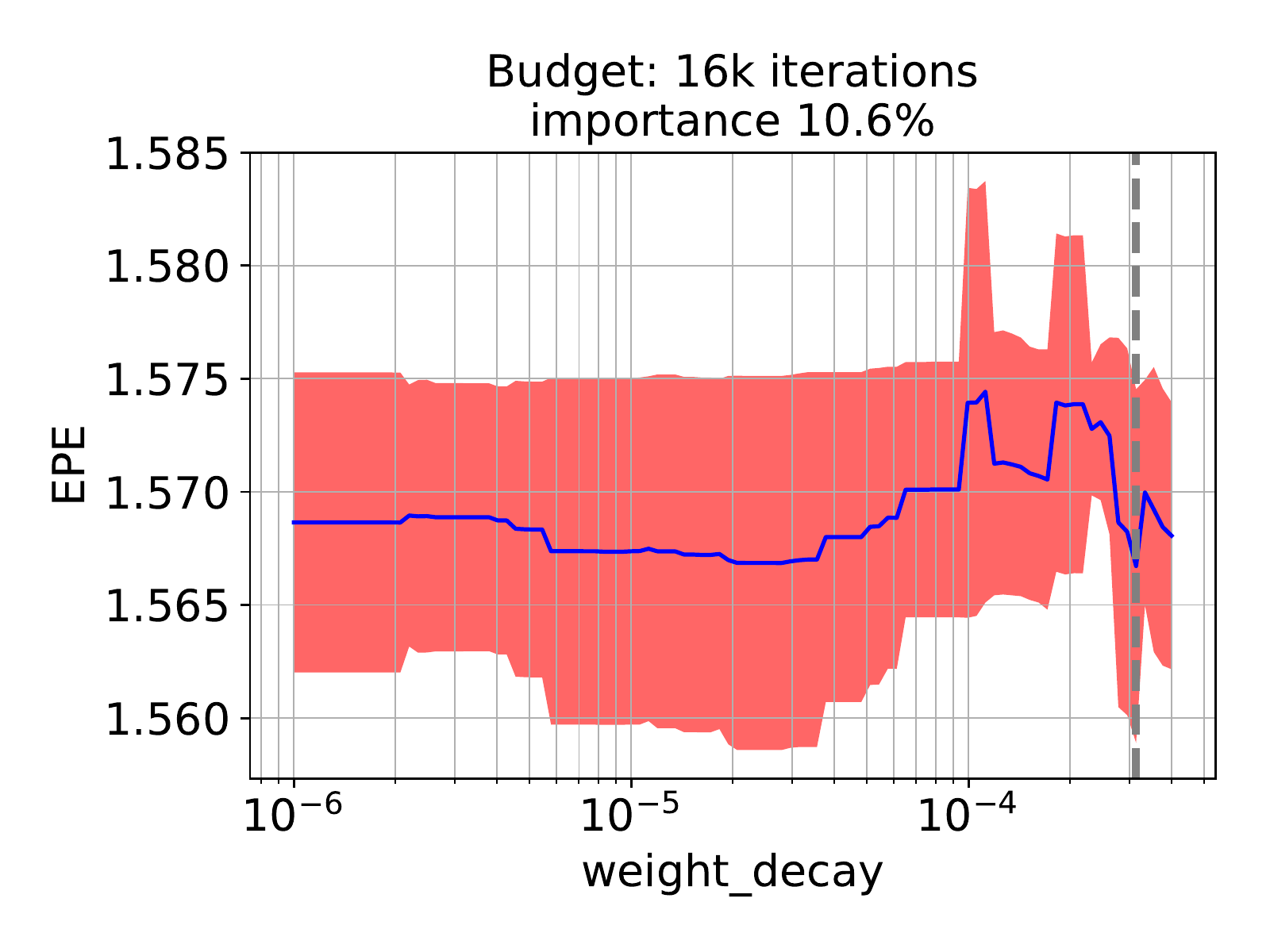}%
%\hfill % <-- Seperation
\includegraphics[width=0.33\textwidth]{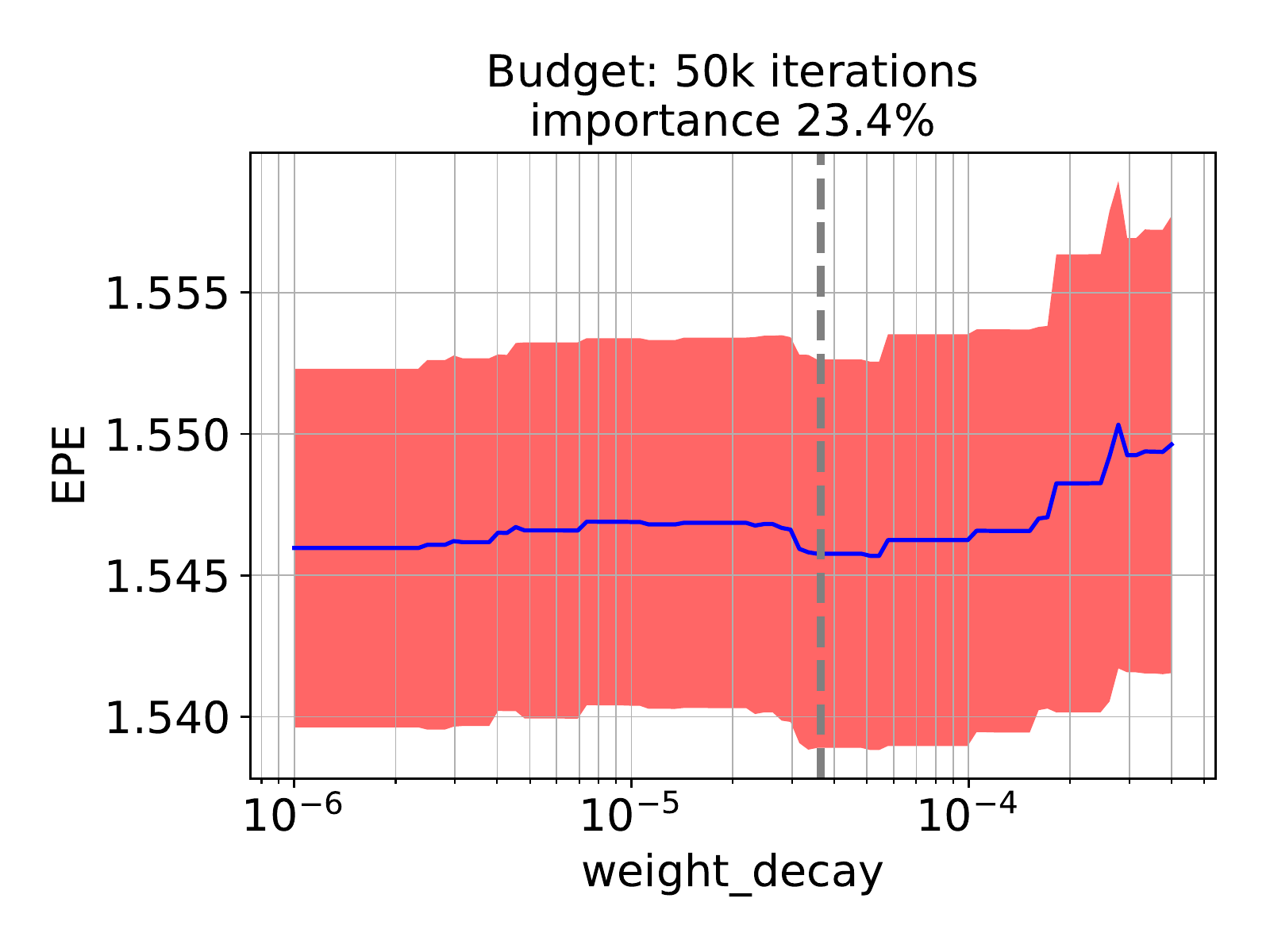}%
%\hfill % <-- Seperation
\includegraphics[width=0.33\textwidth]{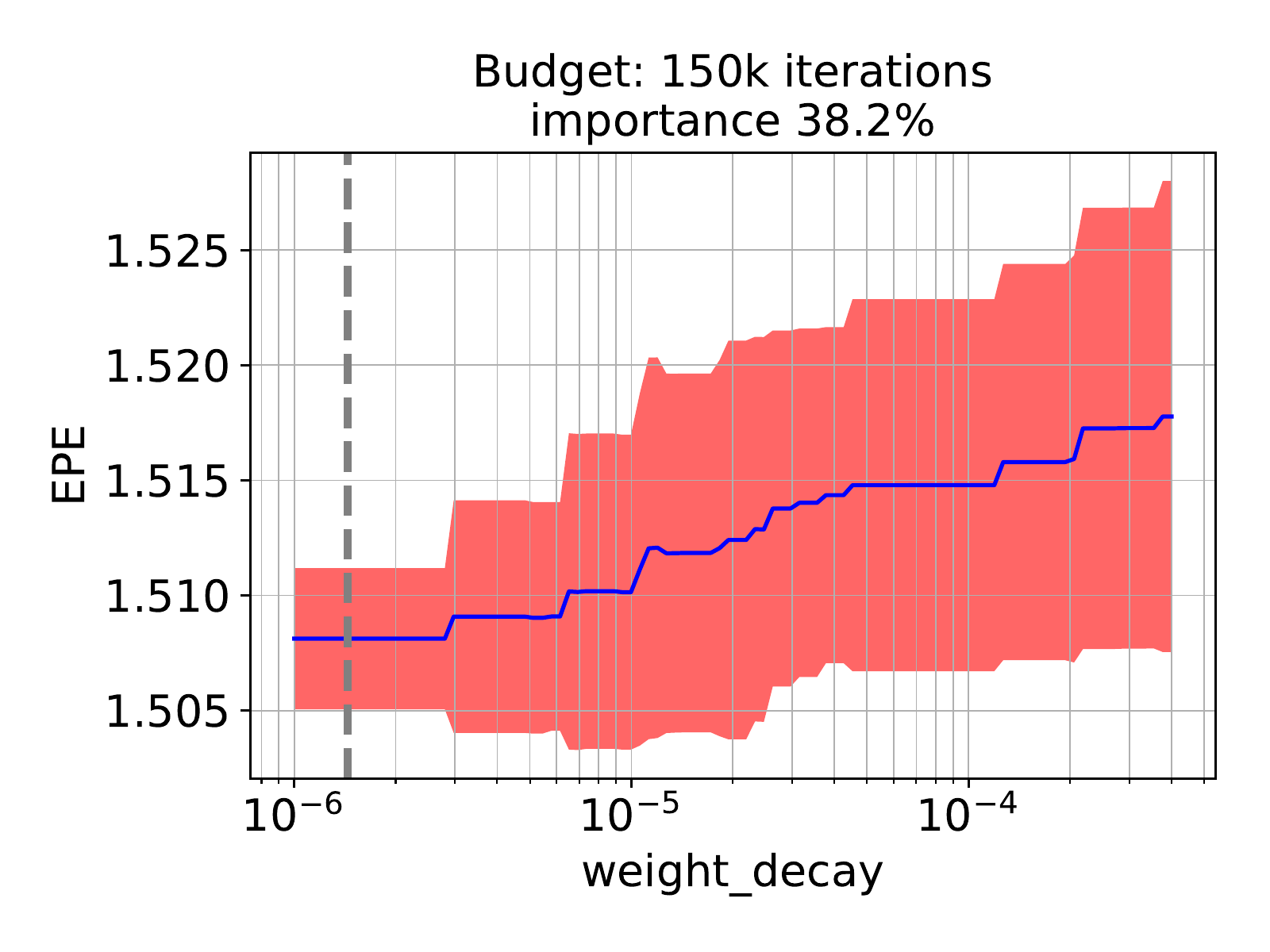}%
\end{center}
\caption{fANOVA plots for all the budgets we run BOHB on the FlyingThings3D dataset. The solid blue line represents the estimated mean EPE (+/- 1std shown by red shaded areas) as a function of hyperparameters as modelled by the random forest we fit to the observations. The importance on top of each plot indicates the fraction of the total variance explained by the individual choice, while the dashed gray line the optimal value as determined by BOHB.}
\label{fig:fanova}
\end{figure*}

\begin{figure*}
\begin{center}
\includegraphics[width=0.33\textwidth]{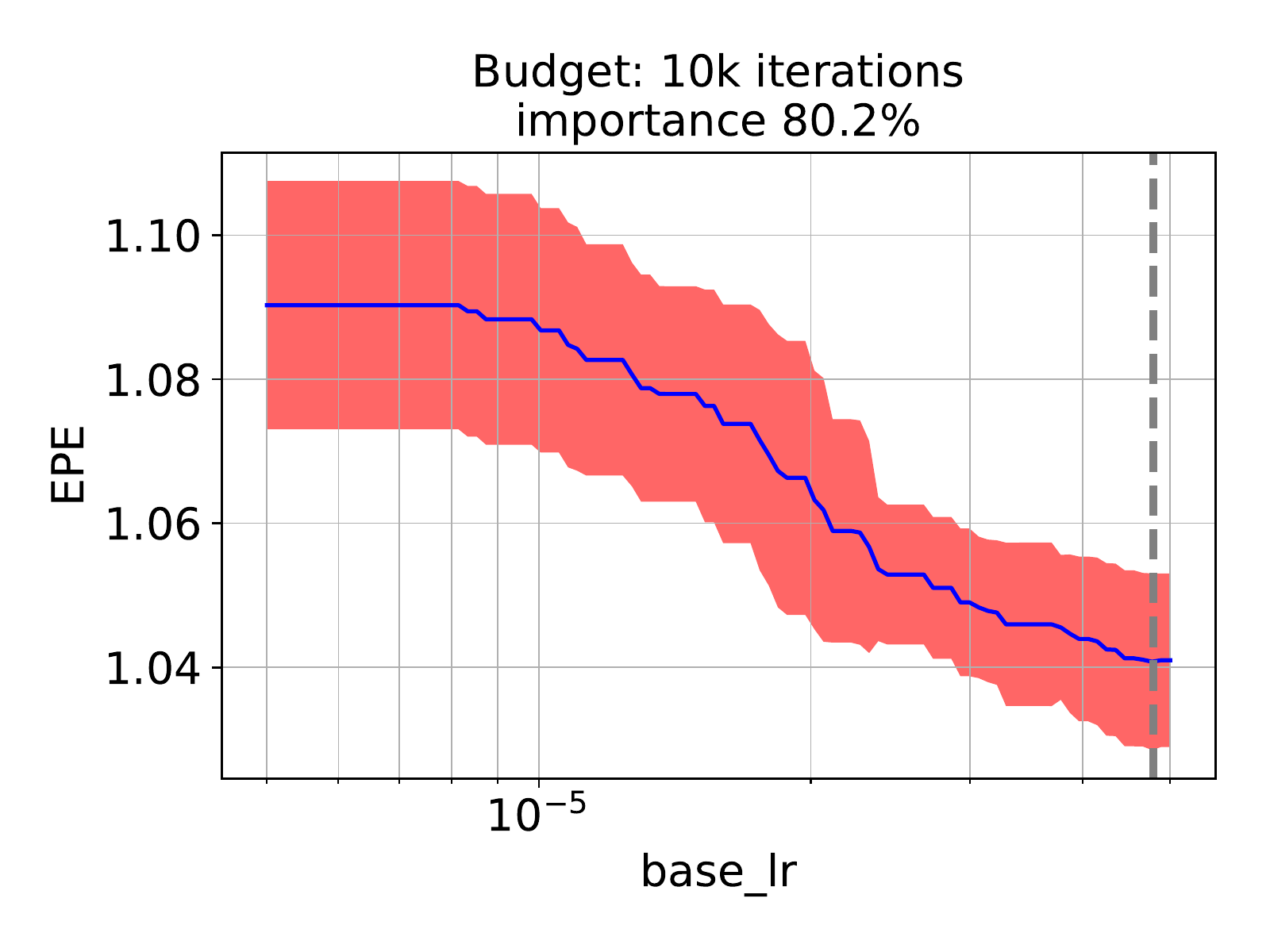}%
%\hfill % <-- Seperation
\includegraphics[width=0.33\textwidth]{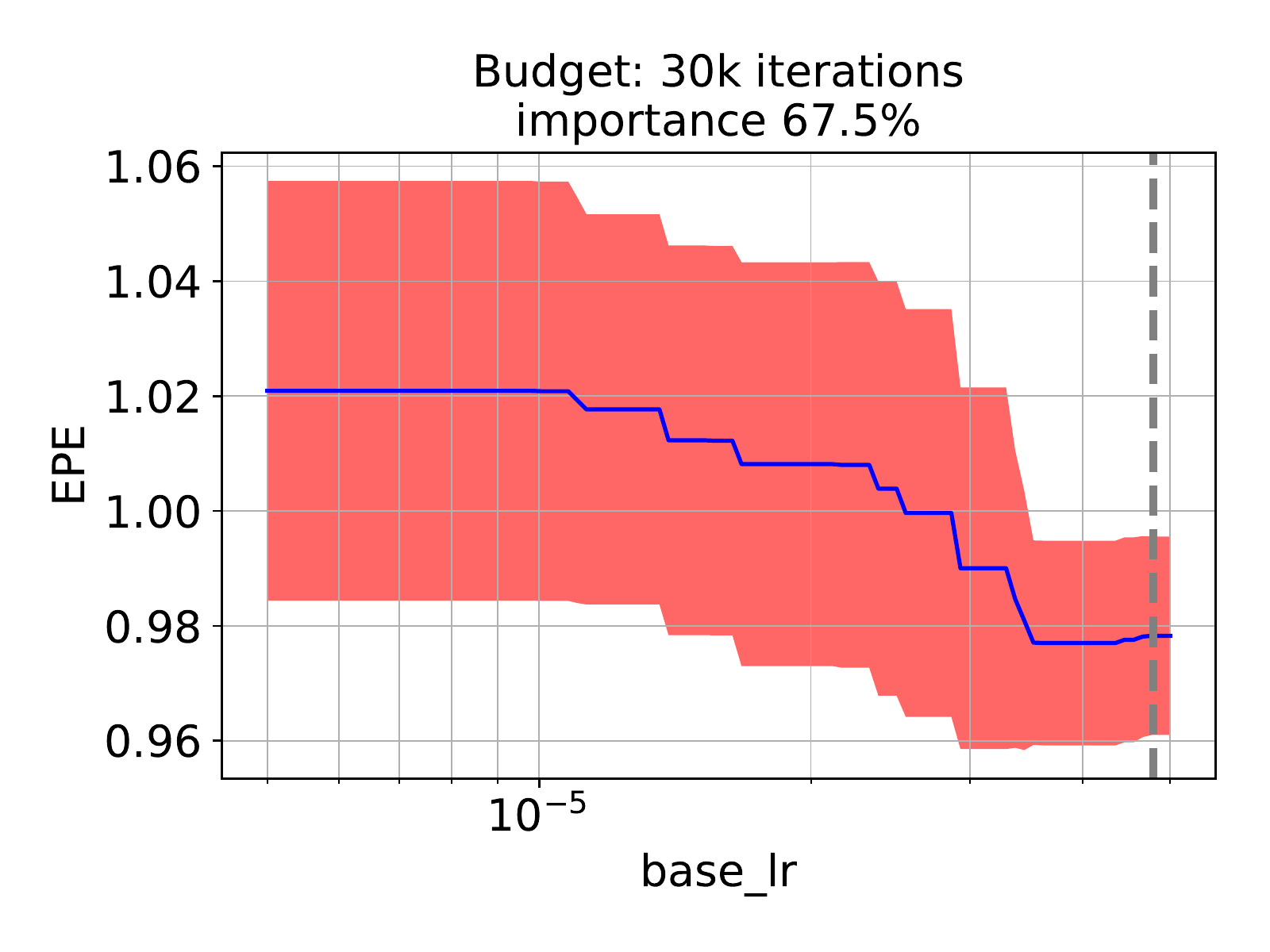}%
%\hfill % <-- Seperation
\includegraphics[width=0.33\textwidth]{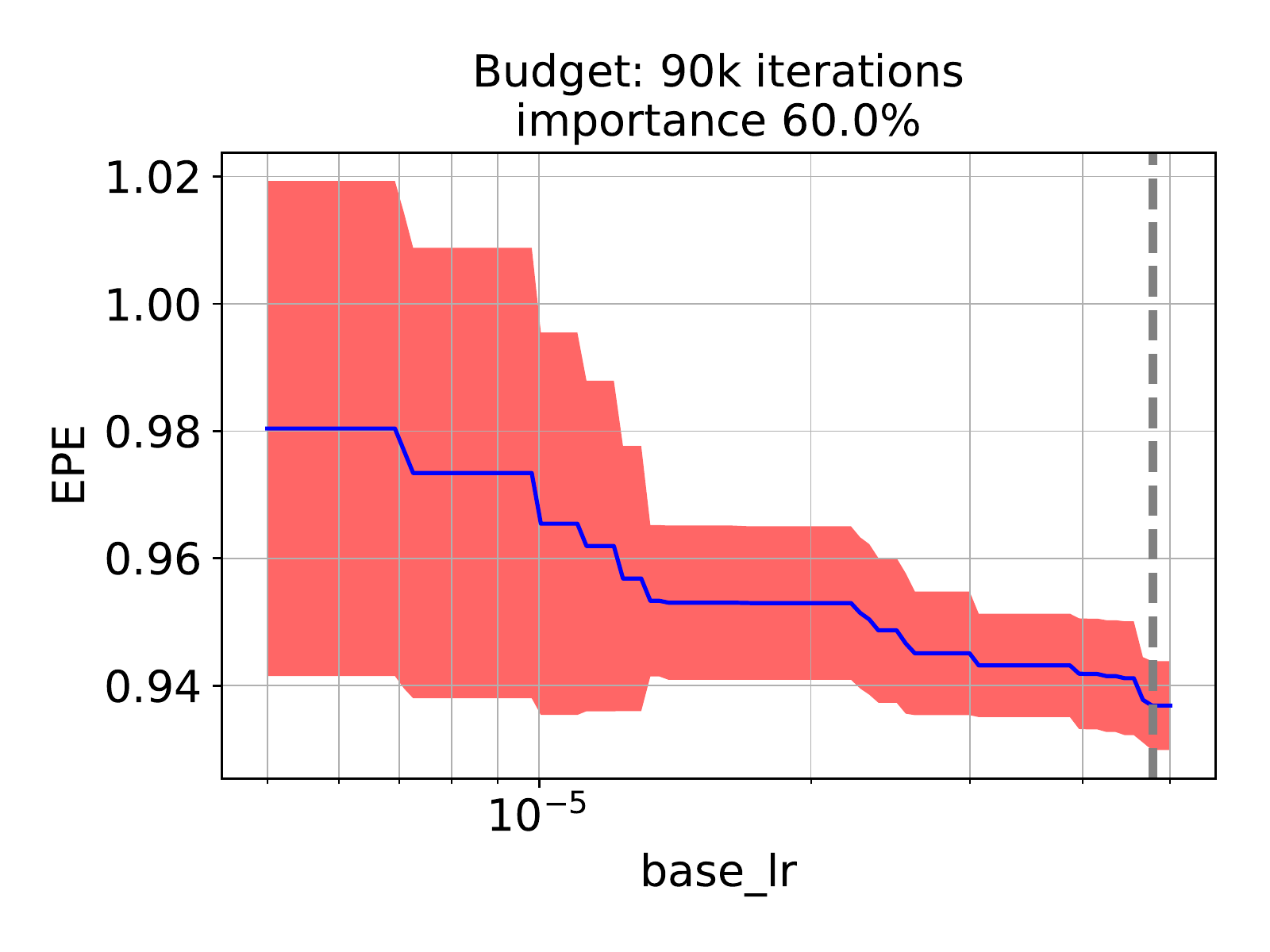}\\%
\includegraphics[width=0.33\textwidth]{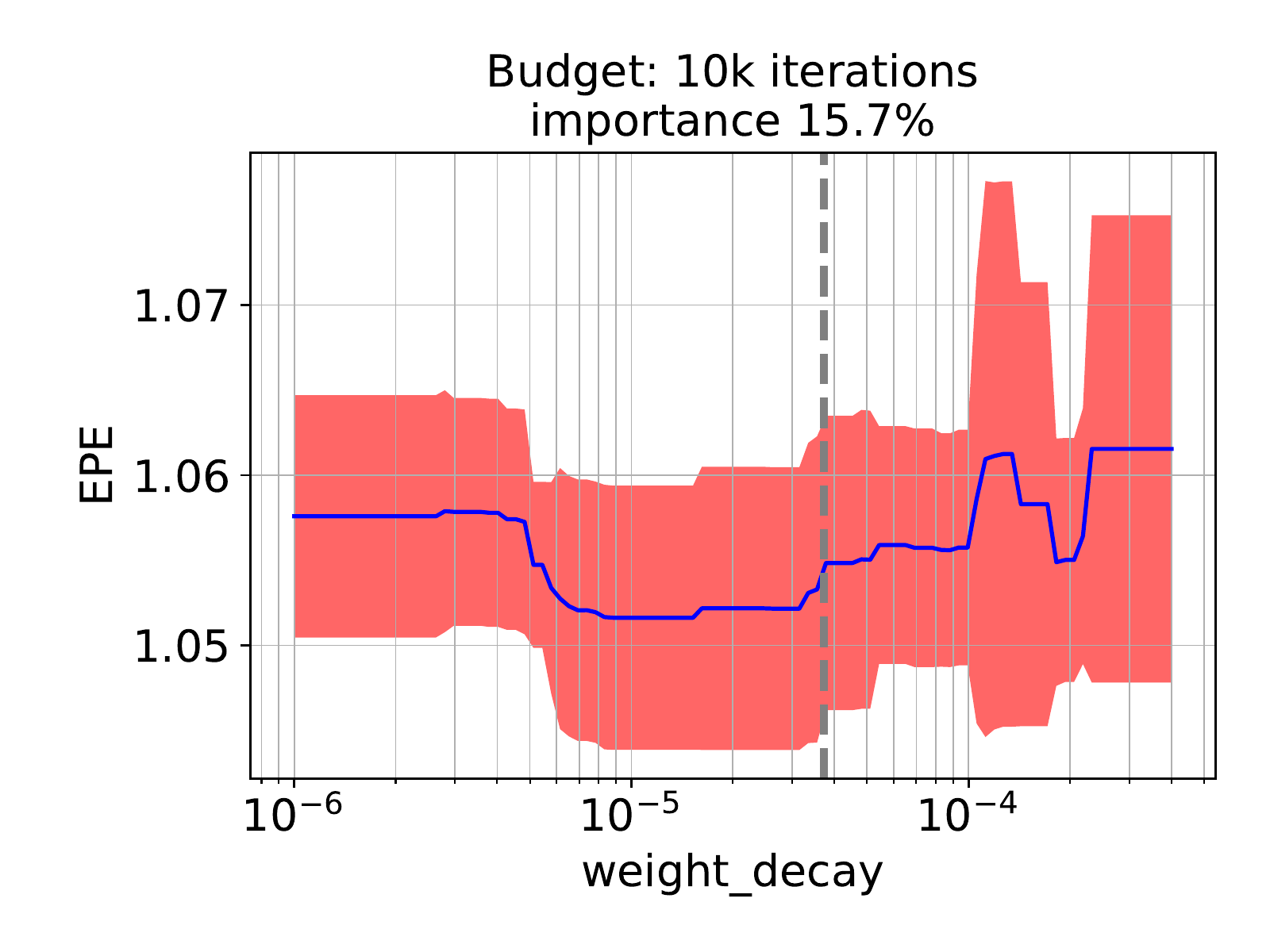}%
%\hfill % <-- Seperation
\includegraphics[width=0.33\textwidth]{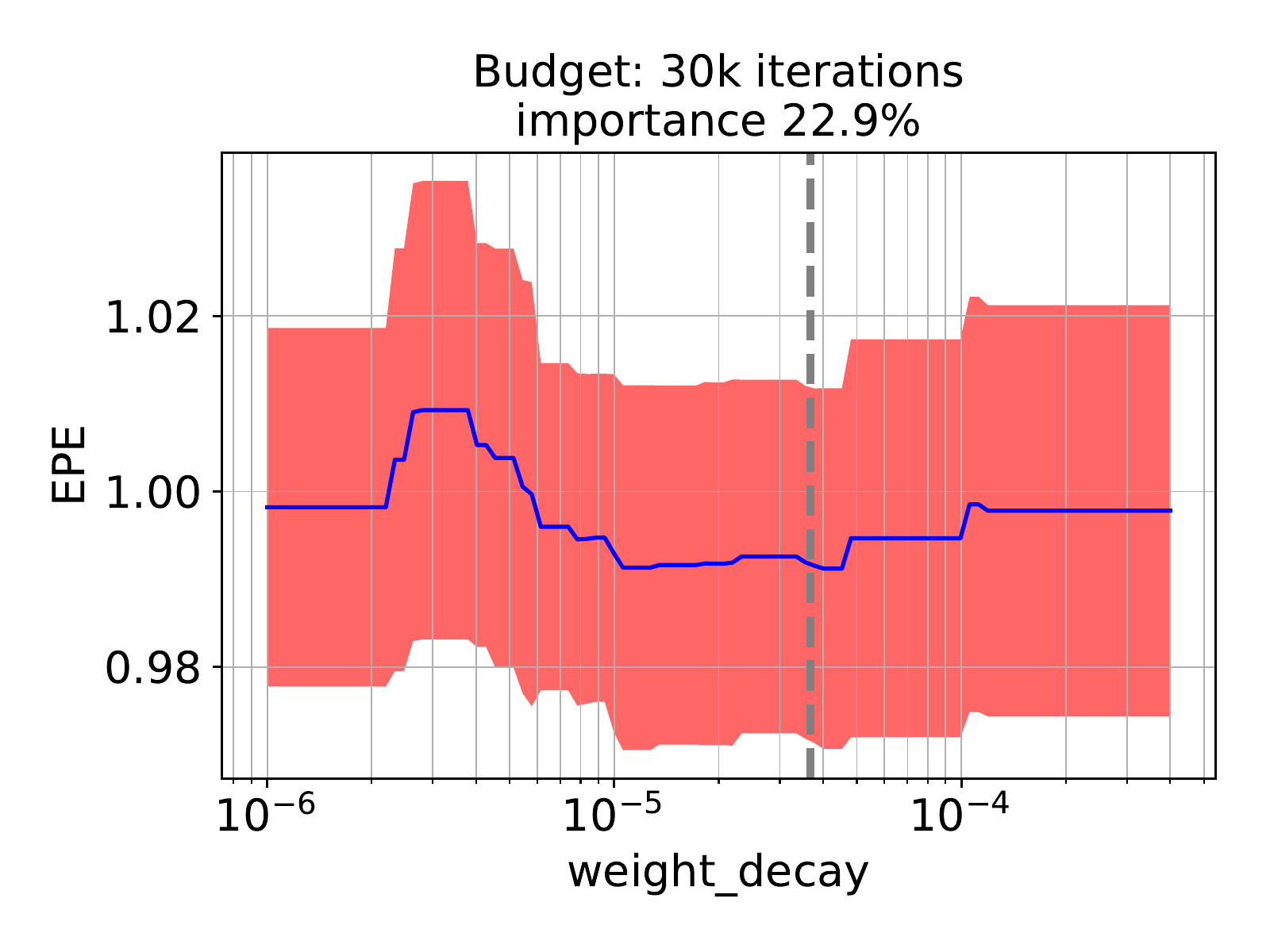}%
%\hfill % <-- Seperation
\includegraphics[width=0.33\textwidth]{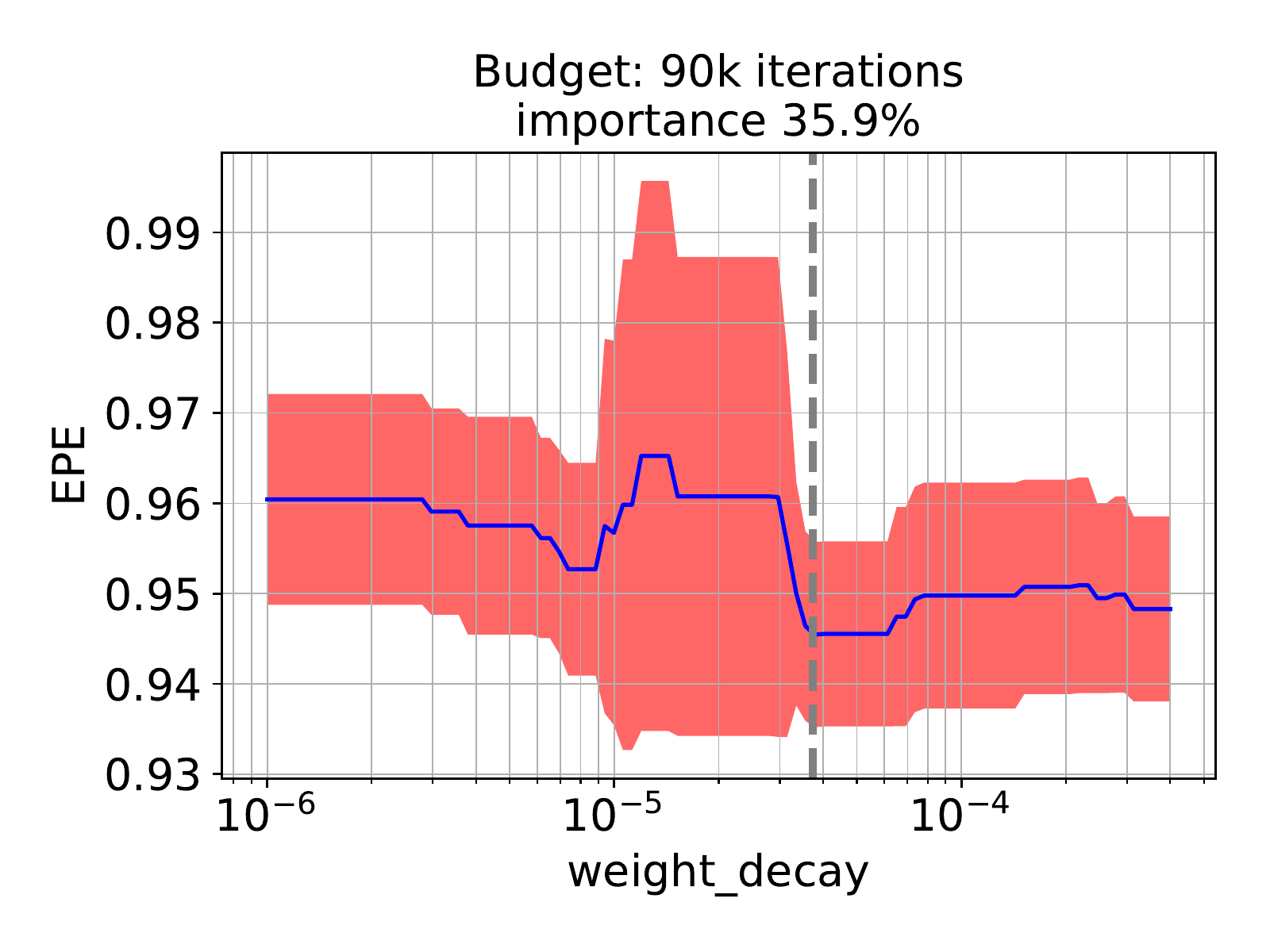}%
\end{center}
\caption{fANOVA plots for all the budgets we run BOHB on the KITTI dataset. The solid blue line represents the estimated mean EPE (+/- 1std shown by red shaded areas) as a function of hyperparameters as modelled by the random forest we fit to the observations. The importance on top of each plot indicates the fraction of the total variance explained by the individual choice, while the dashed gray line the optimal value as determined by BOHB.}
\label{fig:fanova_kitti}

\end{figure*}

\section{Hyperparameter importance}

In order to assess the importance of hyperparameters over the whole search space we analyze our BOHB results using functional analysis of variance (fANOVA; \cite{Hutter14}). This method allows us to quantify how much of the performance variance in the configuration space is explained by single hyperparameters, by marginalizing performances over all possible values that other hyperparameters could have taken. These estimates stem from a random forest model fit on all configurations evaluated on specific budgets during the BOHB optimization procedure.

For the hyperparameter optimization conducted on the FlyingThings3D dataset we observe from \figref{fig:fanova} that the learning rate remains much more important than the weight decay across the first two budgets ($16k$ and $50k$ iterations).
For the highest budget of $150k$ iterations, the importance of the weight decay hyperparameter becomes larger, however it is still dominated by the learning rate. Notice the optimal value that BOHB determines for each hyperparameter in our space (gray dashed line in \figref{fig:fanova}). Interestingly, for smaller budgets (i.e. less training iterations) AutoDispNet-C models trained with a small learning rate and high weight decay value (this has a small importance though) perform better on average. As the budget increases the a higher learning rate and a smaller weight decay value are preferred.

We observe similar results when optimizing the learning rate and weight decay for AutoDispNet-C on the KITTI dataset. From the plots in \figref{fig:fanova_kitti} we can see that the learning rate has a higher contribution to the total performance variance throughout all budgets compared to weight decay. However, the optimal values for these two hyperparameters, as determined by BOHB, remain unchanged across these budgets.

\end{document}